\documentclass{article}

\usepackage[preprint]{neurips_2020}

\usepackage[utf8]{inputenc} % allow utf-8 input
\usepackage[T1]{fontenc}    % use 8-bit T1 fonts
\usepackage{hyperref}       % hyperlinks
\usepackage{url}            % simple URL typesetting
\usepackage{booktabs}       % professional-quality tables
\usepackage{amsfonts}       % blackboard math symbols
\usepackage{nicefrac}       % compact symbols for 1/2, etc.
\usepackage{microtype}      % microtypography

\usepackage{amsmath, amsthm, xcolor, tikz, pgfplots, caption, subcaption, float}
\graphicspath{{figures/}}

\theoremstyle{plain}
\newtheorem{lemma}{Lemma}
\newtheorem{proposition}{Proposition}
\theoremstyle{definition}
\newtheorem*{definition}{Definition}

\title{Scaling laws for single-agent reinforcement learning}

\author{
  Jacob Hilton \\
  OpenAI \\
  \texttt{jacob.hilton@gmail.com} \\
  \And
  Jie Tang \\
  OpenAI \\
  \texttt{jietang@openai.com} \\
  \And
  John Schulman \\
  OpenAI \\
  \texttt{joschu@openai.com} \\
}

\begin{document}

\maketitle

\begin{abstract}
Recent work has shown that, in generative modeling, cross-entropy loss improves smoothly with model size and training compute, following a power law plus constant scaling law. One challenge in extending these results to reinforcement learning is that the main performance objective of interest, mean episode return, need not vary smoothly. To overcome this, we introduce \textit{intrinsic performance}, a monotonic function of the return defined as the minimum compute required to achieve the given return across a family of models of different sizes. We find that, across a range of environments, intrinsic performance scales as a power law in model size and environment interactions. Consequently, as in generative modeling, the optimal model size scales as a power law in the training compute budget. Furthermore, we study how this relationship varies with the environment and with other properties of the training setup. In particular, using a toy MNIST-based environment, we show that varying the ``horizon length'' of the task mostly changes the coefficient but not the exponent of this relationship.
\end{abstract}

\section{Introduction}

Recent studies of how neural network performance varies with model size and training compute have found these relationships to be governed by smooth power laws \citep{languagescalinglaws,scalingcompendium,acousticscalinglaws,translationscalinglaws}. These studies have focused primarily on generative modeling, in which the training objective is cross-entropy loss, and have found test loss to scale smoothly. In this work we seek to extend these results to reinforcement learning, in which there is generally no cross-entropy loss.

In some reinforcement learning environments, there is still a performance metric that varies smoothly \citep{adascaling}. For example, in competitive games, it is often possible to assign Elo ratings to players such that scaled differences in Elo ratings give approximate logit probabilities of victory. Recently it has been shown that, in the board games Hex \citep{hexscalinglaws}, Connect Four and Pentago \citep{connectfourpentagoscalinglaws}, the exponentiated Elo rating of a policy trained using AlphaZero \citep{alphazero} follows a power law in training compute (within a certain Elo range). We call metrics that follow such simple relationships \textit{natural performance metrics}.

However, in other reinforcement learning environments, there may be no obvious natural performance metric. For example, there may be no reason to expect the number of objects collected in a video game to vary smoothly. One approach to overcoming this difficulty is to use ``broken'' power laws \citep{brokenscalinglaws}. As an alternative, we introduce \textit{intrinsic performance}, which is defined to be equal to training compute on the compute-efficient frontier of the tradeoff between model size and environment interactions. This causes the relationship between performance and training compute to follow a power law by definition, thereby making it possible to study the remaining relationships between performance, model size and environment interactions.

We study these relationships across a range of environments: the easy and hard modes of environments from Procgen Benchmark \citep{procgen}; a 1v1 version of Dota 2 \citep{dota}; and a toy environment based on MNIST \citep{mnist} for which we vary the ``horizon length''. Across these environments, we find intrinsic performance to scale as a power law in model size and environment interactions, in much the same way as the analogous quantities in generative modeling.

One consequence of this scaling law is that, as in generative modeling, the optimal model size for a given training compute budget follows a power law. We study in detail how the coefficient and exponent of this relationship vary with properties of the training setup, including: the difficulty mode of environment, for Procgen; the horizon length of the task, for the MNIST-based environment; the period of training used to fit the power law; and whether the width or depth of the model is scaled.

\tableofcontents

\section{Scaling laws without cross-entropy loss}\label{scaling-laws-without-cross-entropy-loss}

\subsection{Intrinsic performance}

In generative modeling, cross-entropy test loss scales smoothly with training compute, following a power law plus constant scaling law \citep{scalingcompendium}. However, in reinforcement learning (RL), there is generally no cross-entropy loss, and the usual objective of mean episode return need not scale so smoothly.

\begin{figure}
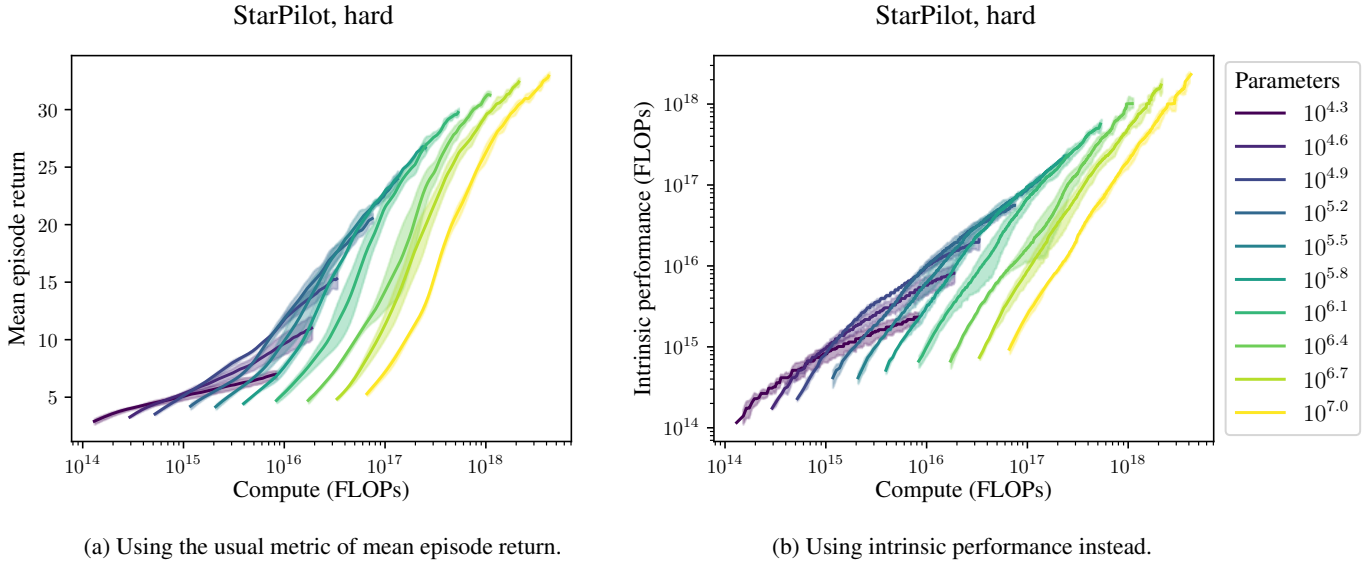

  \begin{changemargin}{-1.0in}{-.75in}
    \begin{subfigure}[t]{0.48\linewidth}
      \centerline{\scalebox{.75}{\input{figures/starpilot_data_return.pgf}}} % PLACEHOLDER_starpilot_data_return.pgf
      \captionsetup{margin={0.175\linewidth,0.075\linewidth}}
      \caption{Using the usual metric of mean episode return.}
      \label{figure-starpilot-data-return}
    \end{subfigure}
    \hspace{.02\linewidth}
    \begin{subfigure}[t]{0.48\linewidth}
      \centerline{\scalebox{.75}{\input{figures/starpilot_data_intrinsic.pgf}}} % PLACEHOLDER_starpilot_data_intrinsic.pgf
      \captionsetup{margin={0.075\linewidth,0.175\linewidth}}
      \caption{Using intrinsic performance instead.}
      \label{figure-starpilot-data-intrinsic}
    \end{subfigure}
  \end{changemargin}
  \caption{Learning curves as a function of total training compute for StarPilot, an environment from Procgen Benchmark, using CNNs of different widths. Mean $\pm 1$ sample standard deviation over three seeds shown.}
  \label{figure-starpilot-data}
\end{figure}

For example, consider StarPilot, a side-scrolling shooter from Procgen Benchmark \citep{procgen}. The agent receives a reward of 1 for destroying each enemy, and the episode continues until either the agent is destroyed, or the agent reaches the end of the level and obtains a bonus reward of 10. There is no reason to expect mean episode return in this game to scale smoothly. Indeed, it takes some ability with aiming and dodging to reach a mean episode return of 5 or 10, but not much additional skill to reach a mean episode return of 15 or 20. This irregular difficulty profile is reflected in the uneven shape of learning curves for this environment (see Figure \ref{figure-starpilot-data}(\subref{figure-starpilot-data-return})).

It may be tempting to conclude that the scaling law methodology cannot be applied to such an environment. However, in generative modeling, there are smooth scaling laws that do not depend on test loss per se. For example, the \textit{model size that achieves} the minimum test loss for a given compute budget scales as a power law with compute. In order to study such relationships in the context of RL, we would like a performance metric that behaves like test loss, i.e., some monotonic function of the return that scales as a power law with compute. We achieve this with our notion of intrinsic performance by simply using \textit{compute itself} as our performance metric.

\begin{definition}
A \textit{scalable model family} is collection of models trained in a uniform way, parameterized by the model size and the total compute used in training. Given a scalable model family, the \textit{intrinsic performance} of an arbitrary policy is the minimum compute required to train a model of any size in the family to reach the same return (averaged over random seeds).
\end{definition}

Another way of explaining this definition is to consider learning curves as a function of compute for a family of models of different sizes, as in Figure \ref{figure-starpilot-data}. The maximum performance over all model sizes defines the \textit{compute-efficient frontier}. When using the usual metric of mean episode return (as in Figure \ref{figure-starpilot-data}(\subref{figure-starpilot-data-return})), the compute-efficient frontier need not follow any particular trend. However, when using intrinsic performance instead (as in Figure \ref{figure-starpilot-data}(\subref{figure-starpilot-data-intrinsic})), the efficient frontier is mapped onto the line $y=x$ by definition. This reveals the regularity of the learning curves, which, as we shall see next, now follow a power law trend.

We describe in detail how we compute intrinsic performance in Appendix \ref{appendix-curve-fitting-methodology}.

\subsection{The power law for intrinsic performance}

Our main empirical result is that intrinsic performance $I$ scales approximately as a power law with model parameters $N$ and environment interactions $E$,
\begin{equation}\label{eq:intrinsic}
\boxed{I^{-\beta}=\left(\frac{N_c}N\right)^{\alpha_N}+\left(\frac{E_c}E\right)^{\alpha_E}},
\end{equation}
where $\alpha_N$, $\alpha_E$, $\beta$, $N_c$ and $E_c$ are positive constants.

This is essentially the same as the corresponding scaling law for language models \citep[equation (1.6)]{languagescalinglaws}, but with test loss replaced by $I^{-\beta}$. Although it appears that we have introduced an additional exponent $\beta$, the intrinsic definition of $I$ means that $\beta$ is actually determined by $\alpha_N$ and $\alpha_E$ (see Lemma \ref{lemma}).

The intuition behind this equation is that, when the number of interactions is not bottlenecked ($E\to\infty$), $I$ scales as a power law in $N$, and when model size is not bottlenecked ($N\to\infty$), $I$ scales as a power law in $E$.

\begin{figure}
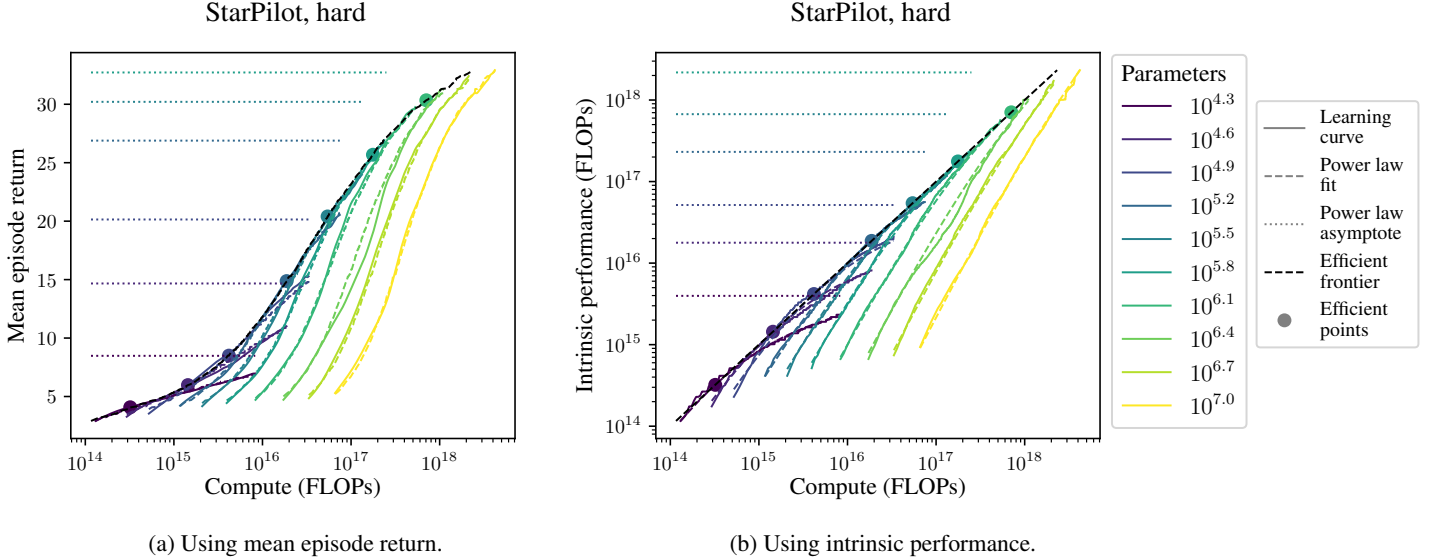

  \begin{changemargin}{-1.3in}{-.75in}
    \begin{subfigure}[t]{0.48\linewidth}
      \centerline{\scalebox{.75}{\input{figures/starpilot_fit_return.pgf}}}
      \captionsetup{margin={0.175\linewidth,0.075\linewidth}}
      \caption{Using mean episode return.}
    \end{subfigure}
    \hspace{.02\linewidth}
    \begin{subfigure}[t]{0.48\linewidth}
      \centerline{\scalebox{.75}{\input{figures/starpilot_fit_intrinsic.pgf}}}
      \captionsetup{margin={0\linewidth,0.325\linewidth}}
      \caption{Using intrinsic performance.}
    \end{subfigure}
  \end{changemargin}
  \caption{Learning curves as a function of total training compute for StarPilot, together with their power law fits. The asymptotes show the $E\to\infty$ limits of the power law fits, representing the predicted performance at convergence. The efficient points show where the power law fits are tangent to the efficient frontier. Mean over three seeds shown.}
  \label{figure-starpilot-fit}
\end{figure}

\subsection{Optimal model size vs compute}\label{optimal-model-size-vs-compute}

An important implication of equation \eqref{eq:intrinsic} is that the optimal model size for a given compute budget scales as a power law in that compute budget.

More precisely, we assume that total training compute is proportional to $NE$ (ignoring the compute required to run the environment, at least for now). Hence, for a given compute budget, there is a trade-off between $N$ and $E$ (the optimum of which defines a point on the compute-efficient frontier). What we will now show is that, under equation \eqref{eq:intrinsic}, the optimal value of $N$ scales as a power law in the compute budget, with an exponent that we will specify.

Since training compute is proportional to $NE$, for convenience we choose units of compute such that training compute equals $NE$ exactly (although in plots we will continue to display compute in FLOPs). This implies that $I=NE$ along the compute-efficient frontier.

\begin{lemma}\label{lemma}
If $I$ satisfies equation \eqref{eq:intrinsic} and $I=NE$ along the compute-efficient frontier, then the compute-efficient frontier is described by the equation
\begin{equation}\label{eq:frontier}
\alpha_N\left(\frac{N_c}N\right)^{\alpha_N}=\alpha_E\left(\frac{E_c}E\right)^{\alpha_E}.
\end{equation}
Moreover, once $\alpha_N$ and $\alpha_E$ are chosen, $\beta$ and $N_cE_c$ are determined:
$$\frac 1\beta=\frac 1{\alpha_N}+\frac 1{\alpha_E}\qquad\text{and}\qquad\frac 1{N_cE_c}=\left(1+\frac{\alpha_N}{\alpha_E}\right)^{\frac 1{\alpha_N}}\left(1+\frac{\alpha_E}{\alpha_N}\right)^{\frac 1{\alpha_E}}.$$
\end{lemma}

For a proof, see Appendix \ref{appendix-proof-of-the-lemma}.

Substituting equation \eqref{eq:frontier} into equation \eqref{eq:intrinsic}, it follows that along the compute-efficient frontier,
$$N=N_c\left(1+\frac{\alpha_N}{\alpha_E}\right)^{\frac 1{\alpha_N}}C^{\frac 1{1+\frac{\alpha_N}{\alpha_E}}},$$
where $C:=NE$. In other words, for a given compute budget $C$, the optimal model size $N$ scales as
$$\boxed{N\propto C^{\frac 1{1+\frac{\alpha_N}{\alpha_E}}}}.$$

\section{Experimental setup}\label{experimental-setup}

\addtocontents{toc}{\protect\setcounter{tocdepth}{1}}

We ran experiments using variety of RL environments:

\begin{itemize}
\item\textbf{Procgen Benchmark} \citep{procgen}: CoinRun, StarPilot and FruitBot in both easy and hard modes, separately varying CNN width and depth.
\item\textbf{Dota 2} \citep{dota}: a 1v1 version of the game, varying LSTM size.
\item\textbf{MNIST}: an RL environment in which the agent has to correctly label a handwritten digit from MNIST \citep{mnist}, using hyperparameters to artificially alter the ``horizon length'' of the task, varying CNN width.
\end{itemize}

All our experiments used a variant of either the PPO algorithm \citep{ppo} or its close cousin PPG \citep{ppg}, along with the Adam optimization algorithm \citep{adam}.

The remainder of this section discusses further details of our experimental setup. Hyperparameters for all our experiments are given in Appendix \ref{appendix-hyperparameters}.

\subsection{Procgen Benchmark}\label{experimental-setup-procgen}

For our Procgen Benchmark experiments, we used CoinRun, StarPilot and FruitBot. We chose these environments because they have lower-variance learning curves than other Procgen environments, and because CoinRun's binary reward enabled us to study the scaling of natural performance metrics (see Section \ref{natural-performance-metrics}). We used both the easy and hard difficulty modes of these environments to see if this would have an effect on the scaling constants.

We used PPG-EWMA \citep{ppoewma} with a fixed KL penalty objective \citep{ppg}, and trained for 200 million environment interactions.

We used the CNN architecture from IMPALA \citep{impala} and conducted both width-scaling and depth-scaling experiments. For our width-scaling experiments, we varied the total number of parameters from $\frac{1}{64}$ of the default to $8$ times the default, rounding to integer numbers of channels. For our depth-scaling experiments, we varied the number of residual blocks per stack from 1 to 64, and used $\frac{1}{4}$ of the default width since the default number of residual blocks per stack was only 2.

\subsection{Dota 2}

For our Dota 2 experiments, we used a 1v1 version of the game to save computational expense.

Following \citet{dota}, we used PPO, but we adjusted the asynchronous setup to ensure that training used only on-policy data with no data reuse. We used 8 parallel GPU workers and trained for between 13.6 billion and 82.6 billion environment interactions.

We used an LSTM architecture and varied the width of the network, with the sizes of the embedding and hidden state varying from 8 to 4096.

\subsection{MNIST}\label{mnist}

Our MNIST environment samples a handwritten digit from the MNIST training set uniformly and independently random at each timestep, and provides an immediate reward of 1 for a correct label and 0 for an incorrect label. There are no episode boundaries, and so we measure mean training accuracy instead of mean episode return.

The use of immediate rewards with no episode boundaries allows the horizon length of the task to be artificially controlled by varying the hyperparameters of our method advantage estimation, GAE \citep{gae}. First, we set the GAE credit assignment parameter $\lambda$ to 1, so that the algorithm assigns credit for each reward to all previous actions, instead of assigning more immediate credit. Then we vary the GAE discount rate $\gamma$, so that the algorithm discounts future rewards at this rate. In separate experiments, we set $\gamma=1-\frac 2{h+1}$ for different values of the ``horizon length'' $h$ ranging from 1 to 256. (This equation is equivalent to saying that an exponentially-weighted moving average with decay parameter $\gamma$ has the same center of mass as the interval $\left[0,h-1\right]$.)

We used PPO-EWMA \citep{ppoewma} with rollouts of length 512 (twice as long as our maximum value of $h$), and trained for $2^{25}$ environment interactions.

We used a simple CNN architecture with ReLU activations and the following layers: a $5\times 5$ convolutional layer with 40 channels, $2\times 2$ max pooling, a $3\times 3$ convolutional layer with 80 channels, $2\times 2$ max pooling, and a dense layer with 1,000 channels. We scaled the width of this network by varying total number of parameters from $\frac{1}{64}$ of the default to $8$ times the default. We used separate policy and value function networks because we did not expect there to be much transfer between the two objectives, since the environment samples digits independently.

\subsection{Learning rates}\label{learning-rates}

Although we would not expect our qualitative results to change much, our quantitative results such as scaling exponents depend crucially on using well-tuned hyperparameters. By far the most important hyperparameter to tune in our setup is the Adam learning rate, whose optimal value can vary substantially with model size and compute budget.

When varying model size, we found that a good heuristic is to keep the Adam learning rate proportional to the initialization scale. For our width-scaling experiments, this means keeping the Adam learning rate proportional to $1/{\sqrt{\text{width}}}$, since we use Kaiming He initialization \citep{heinit}. For our Procgen depth-scaling experiments, which use a residual network, it means keeping the Adam learning rate proportional to $1/{\sqrt{\text{depth}^{\frac 1L}}}$, where $L$ is the number of layers per residual block ($L=2$ in our case), since we use an initialization similar to Fixup initialization \citep{fixup}. For Procgen and MNIST, we tuned the learning rate at one model size and followed this heuristic to select the learning rate for the other model sizes. For Dota 2, we tuned the learning rate separately for each model size, but this amounted to following approximately the same heuristic.

When varying the compute budget for a given model size, it can actually be necessary to use separate training runs for each compute budget, each with its own learning rate schedule, rather than taking different snapshots at different points of the same training run \citep{chinchilla}. Unfortunately, due to the challenge of carefully tuning learning rate schedules for RL and the expense of multiplying the number of training runs, we took the latter approach. To mitigate the impact of this, we found a learning rate schedule that seemed to work well for a variety of compute budgets, which we explain in Appendix \ref{batch-size}. Nevertheless, the values of our scaling exponents should be considered uncertain because of this.

\section{Results}

\addtocontents{toc}{\protect\setcounter{tocdepth}{2}}

Our main result is that our power law for intrinsic performance, equation \eqref{eq:intrinsic}, holds across environments and model sizes, at least after an initial transient period of training (which we discuss in more detail in Section \ref{variability-of-exponents-over-training}). This result is supported by the closeness of the power law fit to our learning curves, as shown in Figure \ref{figure-starpilot-fit} for StarPilot and in Appendix \ref{appendix-results-in-full} for all our environments. Our methodology for fitting this power law is described in Appendix \ref{appendix-curve-fitting-methodology}.

It is interesting to study the sensitivity of the exponents $\alpha_N$ and $\alpha_E$, which govern the scaling behavior of $I$ with $N$ and $E$ (and determine the other exponents of interest). The fitted values of these exponents for the different environments are shown in Figure \ref{figure-exponents}. The numerical values of all of the fitted constants may be found in Appendix \ref{appendix-fitted-constants}.

\begin{figure}
  \centerline{\scalebox{.75}{\input{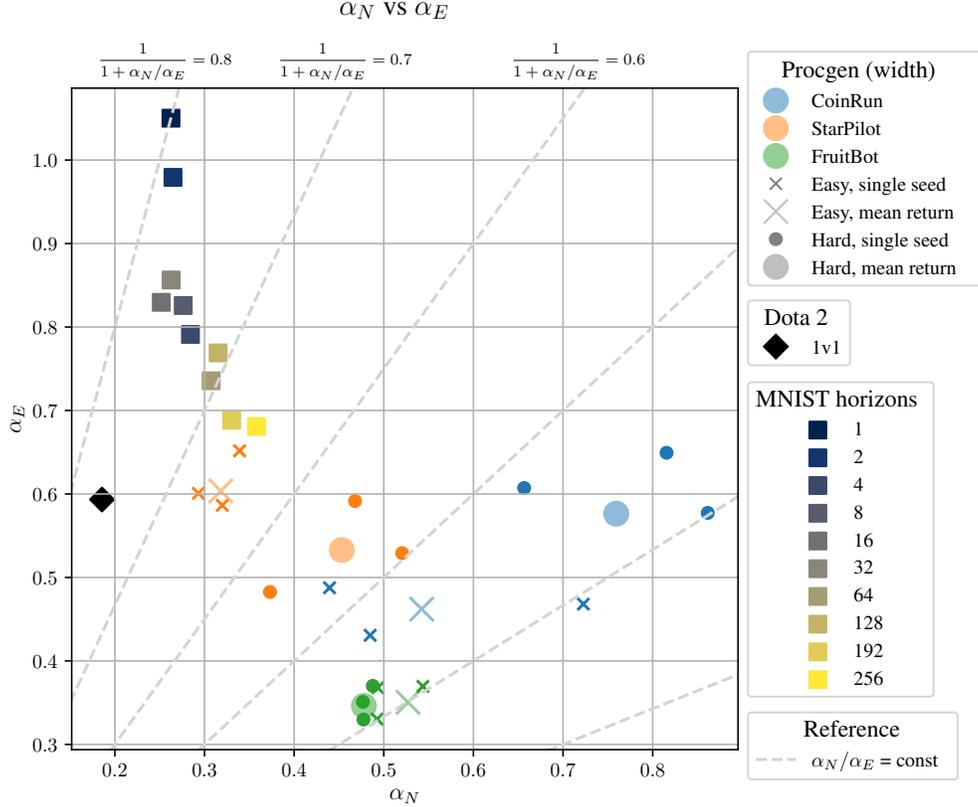}}}
  \caption{Fitted values of $\alpha_N$ and $\alpha_E$. For Procgen, we also show the values fitted using each of the 3 random seeds, to show the variation due to the choice of random seed. The dotted lines show contours for $\frac 1{1+\alpha_N/\alpha_E}$, the exponent for the scaling of optimal model size with compute.}
  \label{figure-exponents}
\end{figure}

Although our measurements of these exponents are uncertain, due to the limitations discussed in Section \ref{limitations}, we make a number of observations:

\begin{itemize}
\item The primary determinant of $\alpha_N$ and $\alpha_E$ is the domain (Procgen, Dota 2, or MNIST), which we expect is a consequence of the fact that so many experimental details are shared within each domain.
\item Within MNIST, increasing the horizon seems to lower $\alpha_E$, but as we explain in Section \ref{effect-of-task-horizon-length}, this effect is confounded by a measurement problem caused by under-training.
\item Within Procgen, the easy and hard modes of each Procgen game tend to have closer exponents to one another than to other Procgen games. We believe that this is because identifying visual features is a core part of Procgen, and the two modes of each game have very similar observation distributions.
\item The Procgen difficulty mode does not obviously have any particular effect on the scaling exponents. We hypothesize that humans tend to judge a task as easier when a near-perfect score can be achieved with less compute, even if it takes a lot of additional compute to eke out the final few points. Conversely, it does not seem to matter to the RL algorithm exactly how the score maps on to intrinsic performance (i.e., the compute required).
\end{itemize}

\subsection{Optimal model size vs compute}

As explained in Section \ref{optimal-model-size-vs-compute}, our power law for intrinsic performance implies that, for a given compute budget, the optimal model size scales as a power law with exponent $\frac 1{1+\alpha_N/\alpha_E}$.

\begin{figure}
  \centerline{\scalebox{.75}{\input{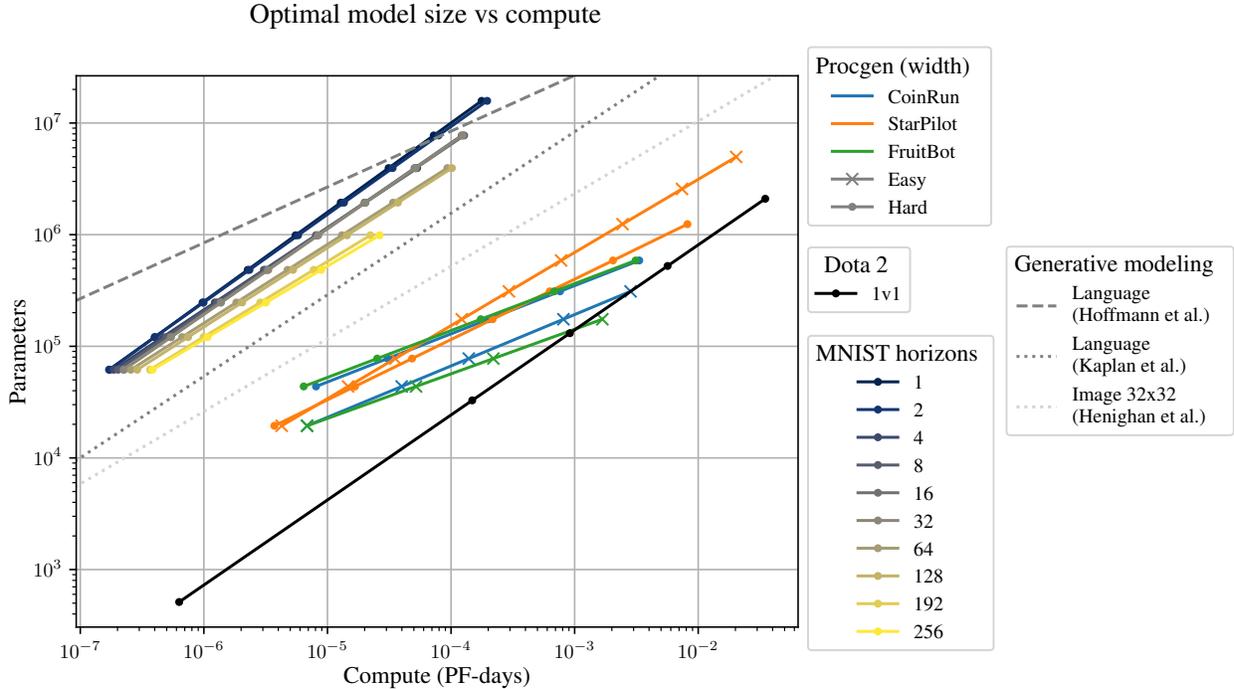}}}
  \caption{Optimal model size vs compute for all our environments. Note that the individual points, which correspond to the sizes of models that we trained, are themselves obtained from a power law best fit. Hence the fact that the lines pass through the points exactly is automatic and does not indicate goodness of fit.}
  \label{figure-optimal-model-size-vs-compute}
\end{figure}

Figure \ref{figure-optimal-model-size-vs-compute} shows these inferred relationships for our different environments, along with some generative modeling relationships taken from the literature. The full equations for these relationships are provided in Appendix \ref{appendix-fitted-constants}.

The exponent $\frac 1{1+\alpha_N/\alpha_E}$ varied between around 0.40 and 0.65 for Procgen and 0.66 and 0.80 for MNIST, and was around 0.76 for Dota 2. By comparison, the corresponding exponent for language modeling, which was carefully measured by \citet{chinchilla}, is around 0.50. Previous work by \citet{languagescalinglaws} and \citet{scalingcompendium} measured this exponent less carefully but using a methodology that more closely matches our own, and found an exponent of around 0.73 for language 0.65 for 32x32 images.

An intriguing conjecture, which is also suggested by theoretical considerations \citep{explainingscaling}, is that the exponent of this relationship would be around 0.5 in every domain if it were measured carefully enough (i.e., with optimal hyperparameters and enough random seeds). Given the limitations of our experiments, we consider our results to be inconclusive on this question.

Nevertheless, it is clear that the scaling \textit{coefficient} of this relationship varies significantly between domains. With the exception of our toy MNIST environment, the optimal model size for RL for a given compute budget is consistently smaller than for generative modeling, in some cases by multiple orders of magnitude. We believe that this is because RL tasks have a longer horizon length than generative modeling in some sense, and explore this hypothesis with our MNIST environment in Section \ref{effect-of-task-horizon-length}. Another possibility is that the arithmetic intensity (i.e., the number of FLOPs per parameter in a forward pass) of the architecture is a confounder, which we discuss in more depth in Section \ref{scaling-depth}.

\subsection{Effect of task horizon length}\label{effect-of-task-horizon-length}

As explained in Section \ref{mnist}, for our MNIST experiments, we artificially altered the ``horizon length'' of the task by setting the GAE credit assignment parameter $\lambda$ to 1 and varying the GAE discount rate $\gamma$.

The expected effect of varying $\gamma$ in this context is given by the following theoretical result.

\begin{proposition}\label{proposition}
Consider an MDP with independent timesteps (by which we mean that each $s_t$ is identically distributed and independent of $s_{t-1}$ and $a_{t-1}$, and episodes never terminate). Suppose we train a model with parameters $\theta$ on this MDP using Vanilla Policy Gradient,\footnote{Vanilla Policy Gradient is a primitive version of PPO, explained here: \url{https://spinningup.openai.com/en/latest/algorithms/vpg.html}} estimating advantages using GAE with $\gamma=1-\frac 2{h+1}$ and $\lambda=1$, and working with separate policy and value function networks. Then the covariance matrix of the policy gradient is approximately
$$\boldsymbol\Sigma_\theta+\boldsymbol\Pi_\theta\left(h+\frac 1h-2\right)$$
for some symmetric positive semi-definite matrices $\boldsymbol\Sigma_\theta$ and $\boldsymbol\Pi_\theta$ that do not depend on $h$.
\end{proposition}

For a proof sketch, see Appendix \ref{appendix-proof-sketch-of-the-proposition}.

Intuitively, this result says that gradient variance may be decomposed into two pieces: one piece that is inherent to the task (the $\boldsymbol\Sigma_\theta$ term), and one piece that comes from imperfect credit assignment (the $\boldsymbol\Pi_\theta$ term). For example, when $h=1$ (i.e., $\gamma=0$), credit is correctly assigned to the previous action only, and hence the second term vanishes. Ignoring the $\frac 1h$ term (since $h\geq 1$), we may stylize this result as: \textbf{gradient variance is an affine function of $h$} (i.e., a linear function with an intercept).

This can be directly translated into a statement about sample efficiency, since multiplying the gradient variance by some factor $c$ can be exactly compensated for by multiplying the batch size by $c$, which multiplies the number of samples used by $c$. Hence in order to reach a given performance level, \textbf{the number of environment interactions required should be an affine function of $h$}. This affine function will come from integrating certain functionals of $\boldsymbol\Sigma_\theta$ and $\boldsymbol\Pi_\theta$ over the course of training, and will therefore depend both on the model architecture and on the choice of performance level.

\begin{figure}
  \centerline{\scalebox{.75}{\input{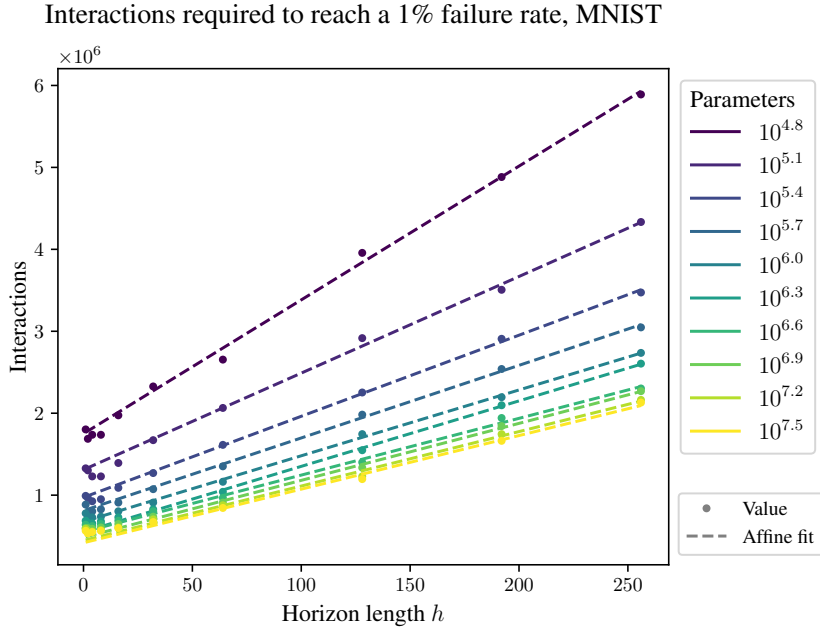}}}
  \caption{Sample efficiency for MNIST as a function of the horizon length $h$, for all our model sizes.}
  \label{figure-mnist-horizon-affine}
\end{figure}

To test this prediction, we looked at the number of environment interactions required to reach a 1\% failure rate (i.e., 99\% training accuracy) on MNIST as a function of the horizon length $h$. Our results are shown in Figure \ref{figure-mnist-horizon-affine}, along with affine fits. As expected, the number of interactions closely follows an affine function of the horizon length, although the fit is less good for shorter horizons and larger models. At very short horizons, the number of interactions even \textit{decreases} with the horizon length, suggesting a hyperparameter issue (perhaps a suboptimal learning rate schedule, or reward normalization implicitly decreasing the KL penalty and entropy bonus).

The implication of this for our optimal model size vs compute scaling law is that once $h$ becomes large enough, further increasing $h$ should lead to a proportional increase the compute budget corresponding to each given optimal model size, without changing the scaling exponent of this relationship. This is because the intercept term of the affine function will eventually become dominated by the term involving $h$, and so the number of environment interactions required to reach a given performance level will eventually scale approximately proportionally to $h$. (For small values of $h$, however, the relationship between the two components of the covariance matrix of the policy gradient may have a more complex dependence on model size.)

This effect is visible in Figure \ref{figure-optimal-model-size-vs-compute}, where the main impact of increasing the horizon length is to shift the optimal model size vs compute curve to the right. The curve also gets shallower as the horizon length is increased, but this effect is confounded by a measurement problem caused by under-training, which we explain in more detail in Section \ref{measurement-problem}.

Our MNIST environment is useful because our it allows us to vary the task horizon length in a fine-grained, quantifiable way by varying $\gamma$. But our analysis of this environment relies on the assumption of independent timesteps, which does not hold in most environments (and in particular removes the need for exploration). Nevertheless, our results are suggestive of a more general explanation for the large differences in optimal model size for a given compute budget between different environments: that different environments have different task horizon lengths in a more general sense. We speculate that, in this more general sense, task horizon length is influenced by how long rewards are delayed for relative to the actions the agent is currently learning (which may increase throughout training as the agent learns skills with feedback loops that are less and less tight), and that $\gamma$ determines only an upper bound on the task horizon length.

\subsection{Variability of exponents over training}\label{variability-of-exponents-over-training}\label{measurement-problem}

Although our power law for intrinsic performance holds across environments and model sizes, we only obtain a good fit by excluding an initial transient period of training. Put another way, the scaling constants vary over the course of training.

This phenomenon is clearest with with our MNIST environment, since we were able to use many random seeds to reduce variance. Recall that in this environment, the agent observes a randomly sampled MNIST training set digit each timestep, and the horizon length of the task is artificially controlled using the GAE discount rate $\gamma$, as explained in Section \ref{mnist}. We fitted our power law to three different periods of training for this environment: an early period ($2^{16}$--$2^{19}$ interactions), a middle period ($2^{19}$--$2^{22}$ interactions), and a late period ($2^{22}$--$2^{25}$ interactions).

\begin{figure}
  \begin{changemargin}{-.375in}{-.375in}
    \centering
    \begin{minipage}[t]{0.49\textwidth}
      \centerline{\scalebox{.75}{\input{figures/mnist_period_exponents.pgf}}}
      \captionof{figure}{Fitted values of $\alpha_N$ and $\alpha_E$ for MNIST with different horizons, using different periods of training to fit the power laws. The horizon $h$ is defined by $\gamma=1-\frac 2{h+1}$, where $\gamma$ is the discount rate.}
      \label{figure-mnist-period-exponents}
    \end{minipage}
    \hspace{0.75in}
    \begin{minipage}[t]{0.49\textwidth}
      \centerline{\scalebox{.75}{\input{figures/mnist_mixed_fit_intrinsic.pgf}}}
      \captionof{figure}{Learning curves as a function of total training compute for MNIST, using different horizons and different periods of training, together with their power law fits. Mean over the middle-performing 16 of 20 random seeds shown.}
      \label{figure-mnist-mixed-fit-intrinsic}
    \end{minipage}
  \end{changemargin}
\end{figure}

Figure \ref{figure-mnist-period-exponents} shows the fitted values of $\alpha_N$ and $\alpha_E$ for these different periods of training. We found $\alpha_E$ to be significantly lower during the early and middle periods of training, especially for the shorter horizon lengths.

In order to accurately measure the scaling constants for optimal model size vs compute, it is best to use a period of training during which the learning curves reach the compute-efficient frontier, since otherwise the measurement is an extrapolation. As shown in Figure \ref{figure-mnist-mixed-fit-intrinsic}, this is always in the late period of training, if at all. For this reason, we use the late period of training for all of our results on MNIST outside of this section.

Figure \ref{figure-mnist-mixed-fit-intrinsic} also shows that, for the longer horizon lengths, the learning curves of the larger models did not reach the compute-efficient frontier even during the late period of training. Hence our measurements of $\frac 1{1+\alpha_N/\alpha_E}$, the exponent for the scaling of optimal model size with compute, are likely underestimates for these longer horizon lengths.

For our other environments, we found that it was enough to exclude only the first $\frac{1}{64}$ of training in order for our power law for intrinsic performance to be a good fit around the compute-efficient frontier. This is similar to what is needed for the corresponding law for language \citep[Figure 4, right]{languagescalinglaws}. Nevertheless, it is possible that the measurement problem identified in this section affects some of our other results.

\subsection{Scaling depth}\label{scaling-depth}

Most of our experiments involved scaling the width of our networks, but for Procgen, we also tried scaling the depth, as explained in Section \ref{experimental-setup-procgen}. We found that our power law for intrinsic performance still held, but with more noise than the width-scaling experiments, as a consequence of using fewer model sizes. The fitted values of $\alpha_N$ and $\alpha_E$ for the depth-scaling experiments lay in a similar region to the width-scaling experiments, but there were no clear relationships between the depth-scaling exponents for the different environments, nor between the width-scaling and depth-scaling exponents for a given environment. Plots of our results may be found in Appendix \ref{appendix-results-in-full}, and the numerical values of the fitted constants may be found in Appendix \ref{appendix-fitted-constants}.

The main difference between our width-scaling and depth-scaling results is that the optimal model size for a given compute budget was significantly smaller for our depth-scaling experiments, but this was an artifact of how we counted parameters and FLOPs. As explained in Appendix \ref{appendix-parameter-and-flop-calculations}, we only included the part of the network being scaled in our parameter and FLOP calculations, which meant excluding the final dense layer of the network for our depth-scaling experiments, but not our width-scaling experiments. If this layer had been included in our depth-scaling calculations, it would have accounted for between 16\% and 90\% of the parameters but only 2\% or fewer of the FLOPs, depending on the depth.

\begin{figure}
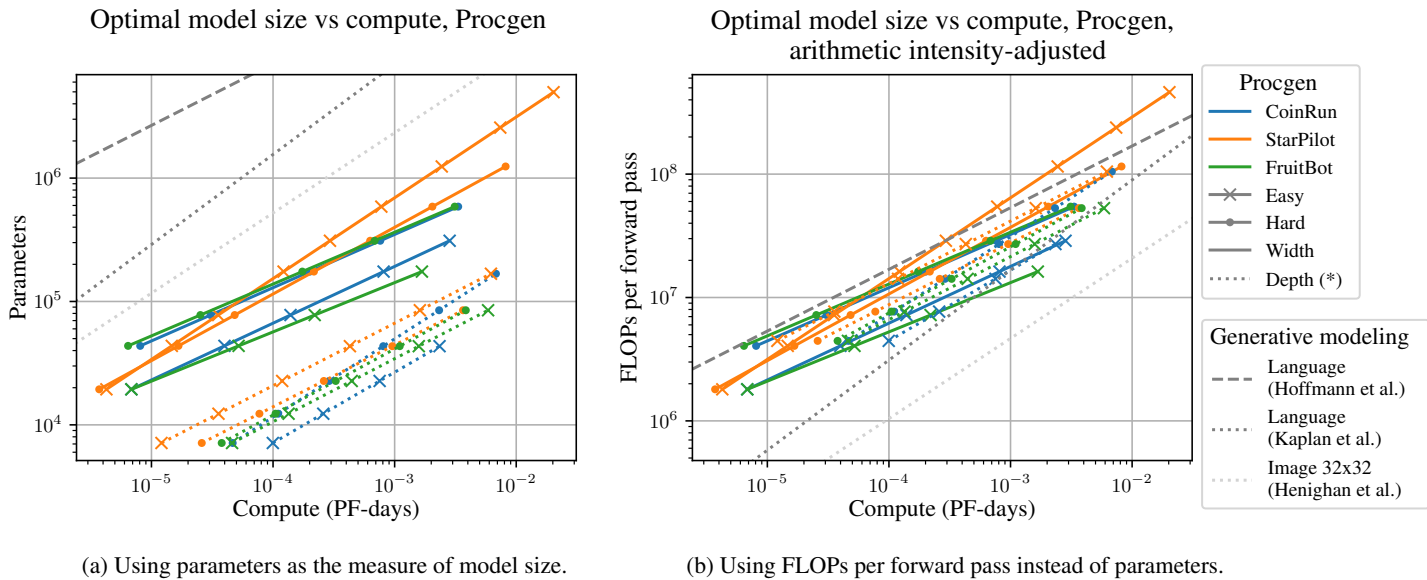

  \begin{changemargin}{-1.3in}{-.75in}
    \begin{subfigure}[t]{0.48\linewidth}
      \centerline{\scalebox{.75}{\input{figures/procgen_optimal_model_size_vs_compute.pgf}}}
      \captionsetup{margin={0.175\linewidth,0.075\linewidth}}
      \caption{Using parameters as the measure of model size.}
    \end{subfigure}
    \hspace{.02\linewidth}
    \begin{subfigure}[t]{0.48\linewidth}
      \centerline{\scalebox{.75}{\input{figures/procgen_optimal_flops_per_forward_pass_vs_compute.pgf}}}
      \captionsetup{margin={0.0125\linewidth,0.2125\linewidth}}
      \caption{Using FLOPs per forward pass instead of parameters.}
    \end{subfigure}
  \end{changemargin}
  \caption{Comparison of optimal model size vs compute for our Procgen width- and depth-scaling experiments. (*) It is important to understand how parameters and FLOPs were counted to interpret the depth-scaling results. This is explained in detail in Appendix \ref{appendix-parameter-and-flop-calculations}.}
  \label{figure-procgen-optimal-model-size-and-flops-per-forward-pass-vs-compute}
\end{figure}

Interestingly, as shown in Figure \ref{figure-procgen-optimal-model-size-and-flops-per-forward-pass-vs-compute}, the optimal model size vs compute scaling laws for our width- and depth-scaling experiments become much more similar if we measure model size using \textit{FLOPs per forward pass} rather than parameters. This is because excluding the final dense layer from the parameter and FLOP calculations significantly increases the arithmetic intensity (i.e., FLOPs per parameter in a forward pass) as calculated for the depth-scaling experiments. This suggests that, when comparing models with very different arithmetic intensities, FLOPs per forward pass may be a better measure of model size than parameters (or perhaps arithmetic intensity should even be considered as an additional independent variable).

\subsection{Natural performance metrics}\label{natural-performance-metrics}

Although in general there may be no obvious performance metric that scales smoothly with model parameters and environment interactions, motivating our use of intrinsic performance, there may still be such a metric in some environments. We call such metrics \textit{natural performance metrics}, and we were able to find them in a couple of our environments:

\begin{itemize}
\item\textbf{CoinRun}: In the CoinRun environment from Procgen Benchmark, the episode return is always either 10 or 0, corresponding to whether or the agent successfully collects the coin at the end of the level. We found the fail-to-success ratio $F:=\frac{10-R}{R}$, where $R$ is the mean episode return, to be a natural performance metric for CoinRun. This is similar to the failure rate $1-\frac{R}{10}$, since $R$ is close to 10 for most of training, but provides a slightly better fit early in training, since it does not have an upper bound of 1. Note that the logarithm of the fail-to-success ratio can also be thought of as the logit function (inverse sigmoid) of the failure rate.
\item\textbf{Dota 2}: Dota 2 is a two-player game, and so the performance of a policy must be measured by comparing it to other policies. The standard method for this is the TrueSkill rating system,\footnote{\url{https://en.wikipedia.org/wiki/TrueSkill}} in which differences in rating between policies correspond to win probabilities when the policies are played against one another, similarly to the Elo rating system. We found TrueSkill to be a natural performance metric for Dota 2.
\end{itemize}

Specifically, we found that our power law for intrinsic performance, equation \eqref{eq:intrinsic}, still roughly held with the left-hand side replaced by a suitable function of the natural performance metric. For CoinRun, we used the fail-to-success ratio directly, but discarded data from early in training where this ratio was above 0.5. For Dota 2, we used $e^{-\alpha_TT}$, where $T$ is TrueSkill and $\alpha_T$ is a fitted constant, which was needed because the scale of $T$ is arbitrary.

Figures \ref{figure-coinrun-fit-frontiers} and \ref{figure-dota-fit-frontiers} compare the efficient frontier fits for intrinsic performance and for the natural performance metric, for CoinRun and Dota 2 respectively. The fits match closely, except for Dota 2 at higher levels of TrueSkill. We conjecture that Dota 2 has an analog of an \textit{irreducible loss} \citep{scalingcompendium}, representing the maximum attainable TrueSkill for the family of models we trained.

\begin{figure}
  \begin{changemargin}{-.5in}{-.5in}
    \centering
    \begin{minipage}[t]{0.49\textwidth}
      \centerline{\scalebox{.75}{\input{figures/coinrun_fit_frontiers.pgf}}}
      \captionof{figure}{Comparison of the efficient frontier fits for CoinRun, using intrinsic performance and the fail-to-success ratio.}
      \label{figure-coinrun-fit-frontiers}
    \end{minipage}
    \hspace{1in}
    \begin{minipage}[t]{0.49\textwidth}
      \centerline{\scalebox{.75}{\input{figures/dota_fit_frontiers.pgf}}}
      \captionof{figure}{Comparison of the efficient frontier fits for Dota 2, using intrinsic performance and exponentiated scaled TrueSkill.}
      \label{figure-dota-fit-frontiers}
    \end{minipage}
  \end{changemargin}
\end{figure}

We explored introducing an additional fitted constant $T^\ast$ for this maximum attainable TrueSkill, and using either of the functional forms $e^{-\alpha_TT}-e^{-\alpha_TT^\ast}$ and $\left(T^\ast-T\right)^{\alpha_T}$. However, it was unclear to us which of these forms made the most theoretical sense, and we were unsure whether we could justify the extra degree of freedom given the lack of data at higher levels of TrueSkill.

The fitted constants for all of these alternative power laws for both CoinRun and Dota 2 are given in Appendix \ref{appendix-fitted-constants}. Interestingly, for CoinRun, the values of the scaling exponent for the fail-to-success ratio $F$ in terms of intrinsic performance $I$, corresponding to the slopes of the lines in Figure \ref{figure-coinrun-fit-frontiers}, are similar between the two difficulty modes: $F\propto I^{-0.40}$ in easy mode and $F\propto I^{-0.48}$ in hard mode.

\section{Discussion}\label{discussion}

\subsection{Extrapolating sample efficiency}\label{extrapolating-sample-efficiency}

We may use our power law for intrinsic performance, equation \eqref{eq:intrinsic}, to extrapolate sample efficiency to unseen model sizes $N$ and environment interactions $E$. For example, in Figure \ref{figure-starpilot-sample-efficiency}, we show the extrapolated learning curve for StarPilot in the infinite-width limit. This reaches the final performance of our largest model in about half the number of environment interactions. Note, however, that without a natural performance metric, we cannot extrapolate to unseen performance levels.

\begin{figure}
  \begin{changemargin}{-.5in}{-.5in}
    \centering
    \begin{minipage}[t]{0.49\textwidth}
      \centerline{\scalebox{.75}{\input{figures/starpilot_sample_efficiency.pgf}}}
      \captionof{figure}{Learning curves for StarPilot (hard mode, scaling width), together with their power law fits, and the $N\to\infty$ limit of the power law.}
      \label{figure-starpilot-sample-efficiency}
    \end{minipage}
    \hspace{1in}
    \begin{minipage}[t]{0.49\textwidth}
      \centerline{\scalebox{.75}{\input{figures/optimal_model_size_vs_compute_environment.pgf}}}
      \captionof{figure}{Optimal model size vs compute, taking into account a hypothetical compute cost per environment interaction equal to that of a model of size $N_e=10^5$. See Figure \ref{figure-optimal-model-size-vs-compute} for the full legend.}
      \label{figure-optimal-model-size-vs-compute-environment}
    \end{minipage}
  \end{changemargin}
\end{figure}

It is natural to ask how this extrapolated infinite-width limit compares to human sample efficiency. On StarPilot (slowed down to 3 frames per second), a human can reach a mean episode return of around 20 after a few episodes, whereas the extrapolated infinitely-wide model takes 18 million interactions, around 10,000 times as many. This is not really a fair comparison though, because much of the challenge in Procgen is to learn to identify basic visual features, which humans are already able to do. For Dota 2, we crudely estimate that it would take a human around 50--500 hours of gameplay to reach the performance of the extrapolated infinitely-wide LSTM after 5 billion interactions, a factor of 100--1,000 in sample efficiency. This comparison may be fairer, because Dota 2 has a structured observation space and is more challenging than StarPilot, although it still draws on many pre-existing human intuitions. Of course, our models were all trained from scratch, and we should expect this factor to be smaller for models that have been pre-trained to learn useful representations.

\subsection{Cost-efficient reinforcement learning}\label{compute-efficient-rl}

In the reinforcement learning literature, sample efficiency is usually taken to be the primary metric of algorithmic progress. This can be thought of as focusing on the cost of running the environment, but not the algorithm. At the other extreme, we have so far focused on the computational cost of the algorithm, but not on the cost of the environment. However, it is straightforward to now take both into account. To do this, let $N_e$ be the cost of the environment, measured in terms of the number of parameters in a model with the same cost per interaction. Thus the total cost of both the algorithm and the environment is proportional to $\left(N+N_e\right)E$.

The cost-efficient frontier is now described by the following generalization of equation \eqref{eq:frontier}:
$$\left(1+\frac{N_e}N\right)\alpha_N\left(\frac{N_c}N\right)^{\alpha_N}=\alpha_E\left(\frac{E_c}E\right)^{\alpha_E}.$$
Substituting this into our power law given by equation \eqref{eq:intrinsic}, it follows that along the cost-efficient frontier,
$$C=\left(1+\frac{N_e}N\right)\left(\frac 1{1+\frac{\alpha_N}{\alpha_E}\left(1+\frac{N_e}N\right)}\right)^{\frac 1{\alpha_N}+\frac 1{\alpha_E}}\left(\frac N{N_c}\right)^{1+\frac{\alpha_N}{\alpha_E}},$$
where $C:=\left(N+N_e\right)E$. Thus for a given budget $C$, the optimal model size $N$ scales as the same power law in $C$ as before once $N\gg N_e$, and it is only efficient to take $N\ll N_e$ when $C$ is very small. This validates and makes precise the rule-of-thumb that it is usually inefficient to use a model that is much cheaper to run than the environment, at least when training from scratch.

To illustrate this relationship, Figure \ref{figure-optimal-model-size-vs-compute-environment} shows the optimal model size vs compute relationship from Figure \ref{figure-optimal-model-size-vs-compute}, but incorporating a fixed hypothetical compute cost associated with each environment interaction.

\subsection{Limitations}\label{limitations}

Our experiments have several limitations:

\begin{itemize}
\item As explained in Section \ref{learning-rates}, we did not use separate training runs for each compute budget, each with their own learning rate schedule, which can be necessary to accurately measure scaling exponents \citep{chinchilla}. We tried to mitigate this by using a learning rate schedule that worked well for a variety of compute budgets, as explained in Appendix \ref{batch-size}, but this may not have been enough.
\item As explained in Section \ref{measurement-problem}, the variability of exponents over training gives rise to a measurement problem. We mitigated this to some extent by excluding data from early in training when fitting our power law, but this does not fully correct for the fact that some of our models were under-trained relative to the compute-efficient frontier.
\item We did not carefully optimize the aspect ratios of our models, instead scaling width and depth separately. More generally, suboptimal hyperparameters or other problems with our training setups could have lead to errors in our measurements of scaling constants.
\item Learning curves in reinforcement learning are often very high-variance, adding significant noise to power law fits. We mitigated this to some extent by choosing environments with relatively low-variance learning curves and using multiple random seeds, but a lot of variance still remained.
\end{itemize}

As a result of these limitations, we do not think conclusions that depend on the precise fitted values of our scaling constants can be drawn with confidence, although we consider our mitigations sufficient for more qualitative conclusions. We are excited for future work to fix these limitations, explore new domains, and more carefully disentangle the effects of the choice of algorithm, architecture and hyperparameters as well as properties of the environment.

\subsection{Forecasting compute requirements}\label{forecasting-compute-requirements}

The scaling of optimal model size with compute is a key input into the biological anchors framework for forecasting transformative artificial intelligence \citep{agitimelines}. In this framework, the human brain is used as a biological anchor for estimating the number of parameters in a transformative model, and optimal model size vs compute scaling laws are used to forecast the total compute required to train such a model. In this section we summarize the main implications of our work for this framework.

\textbf{Scaling exponents for reinforcement learning lie in a similar range to generative modeling.} The exponent for the scaling of optimal model size with compute, $\frac 1{1+\alpha_N/\alpha_E}$, varied between around 0.4 and 0.8 for our environments, a range that encompasses previous measurements of this exponent for generative modeling. However, as discussed in Section \ref{limitations}, we do not think our measurements of this exponent should be taken literally, due to the limitations of our experiments. The results of \citet{chinchilla} and \citet{explainingscaling} suggest the possibility that this exponent would be around 0.5 in every domain if it were measured carefully enough, and we consider our results to be inconclusive on this question.

\textbf{Scaling coefficients for reinforcement learning vary by multiple orders of magnitude.} The coefficient for the scaling of optimal model size with compute, $N_c\left(1+\frac{\alpha_N}{\alpha_E}\right)^{\frac 1{\alpha_N}}$, varied substantially, enough that we do not think this variation is attributable only to the limitations of our experiments. For example, the scaling exponents for MNIST (with a horizon length of 1) and Dota 2 are very similar, but a model of the same size needs to be trained for around 2,000 times longer on Dota 2 than on MNIST to be compute-efficient. By comparison, \citet{scalingcompendium} found generative modeling to require around 20 times as much training on 32x32 images than on language. Moreover, our analysis of the effect of the task horizon length gives a plausible mechanism for this variation.

\textbf{Arithmetic intensity may confound scaling coefficients.} As discussed in Section \ref{scaling-depth}, the coefficient for the scaling of optimal model size with compute can be affected by the arithmetic intensity (i.e., the number of FLOPs per parameter in a forward pass) of the model. This alone does not explain the large variation in this coefficient between MNIST and Dota 2, for example, but it may explain some of the other variation. We hypothesize that, when comparing models with very different arithmetic intensities, due to parameter sharing or methods such as mixture of experts, it may be better to measure model size in \textit{FLOPs per forward pass} rather than in parameters.

\textbf{Number of samples required is an affine function of the task horizon length.} We study the effect of the task horizon length using a toy MNIST-based environment in Section \ref{effect-of-task-horizon-length}. Both theoretically (see Proposition \ref{proposition}) and empirically, the number of samples required to reach a given level of performance grows with the horizon length as an affine function (i.e., a linear function with an intercept) that depends on both the model size and the target performance level. However, our analysis makes a simplifying assumption of independent timesteps, which does not hold in most environments. In particular, we do not analyze the need for curricula and/or exploration to solve tasks for which it is challenging to obtain useful feedback. Instead, we simply assume that the algorithm pays attention to rewards over a longer time horizon, making credit assignment harder.

This result validates and refines the analysis of \citet{agitimelines}, who defined the ``effective horizon length'' as a quantity that scales linearly with training data requirements, incorporating not only the horizon length as we define it, but also reward sparsity, noise and so on. Our result specifically isolates the explicit horizon length, showing that training data requirements are a sum of two components, at least in our toy setting: one corresponding to a version of the task in which the horizon ends immediately, and another that is proportional to the horizon length. This implies that, for a given fixed task, continuing to increase the horizon length will \textit{eventually} lead to a proportional increase in the compute budget corresponding to a given optimal model size, without changing the exponent of this scaling law. However, this will only happen once the first component has become negligible, and it is unclear whether there are realistic tasks of different horizon lengths for which this first component is negligible in practice.

We are excited for future work to study other aspects of the ``effective horizon length'', such as reward sparsity and noise, as well as studying the explicit horizon length in environments that are less artificial. It is not entirely clear how to quantify these properties in general, and they could potentially affect scaling exponents as well as scaling coefficients, if for example they change over the course of training.

\textbf{Measuring scaling exponents precisely is challenging.} The biological anchors framework uses the scaling of optimal model size with compute to perform a substantial extrapolation, making it particularly sensitive to the exponent of this relationship. This makes it challenging to measure this exponent with sufficient precision. In addition to the challenges raised by \citet{chinchilla} involving learning rate schedules, we hope that others will benefit from learning about the other challenges we faced, which are summarized in Section \ref{limitations}.

\section{Conclusion}

We have shown how to extend scaling laws to single-agent reinforcement learning using the notion of \textit{intrinsic performance}. Across a range of environments, intrinsic performance scales as a power law in model size and environment interactions, and hence the optimal model size scales as a power law in the training compute budget. We have studied how this relationship is affected by various properties of the training setup, including the horizon length of the task, and have discussed the implications of this for the biological anchors framework for forecasting transformative artificial intelligence.

\addtocontents{toc}{\protect\setcounter{tocdepth}{0}}

\section{Acknowledgments}

Thanks to Mira Murati, Karl Cobbe, Chris Hesse, David Farhi, Paul Christiano, Jared Kaplan, Long Ouyang and Ajeya Cotra for discussions, ideas, help, advice, support and inspiration that have greatly benefited this project.

\bibliographystyle{abbrvnat}
\bibliography{bibliography}

\appendix

\addtocontents{toc}{\protect\setcounter{tocdepth}{1}}

\newpage

\section{Curve-fitting methodology}\label{appendix-curve-fitting-methodology}

In this section we discuss our methodology for computing intrinsic performance and fitting the power law constants, which require some care. Code for our full procedure, along with its application to our experiments, may be found in this Colab notebook: \url{https://colab.research.google.com/drive/1PzwZyXsi9jRdVCj1GJrS8JdOPBQ7LHZV}.

Recall that the intrinsic performance of a policy is the minimum compute required to train a model of any size in the same family to reach the same return (averaged over random seeds). The naive way to compute this would be to train models of many different sizes, and to take the best-performing model size for each possible compute budget. However, it may not be feasible to train models of enough different sizes to get a reasonable level of granularity, while using enough different random seeds sufficiently to reduce the high variance of learning curves.

To cope with this, we compute intrinsic performance and fit the power law constants together. This allows us to make use of all the data from each learning curve, instead of just a single point from each one. We do this by jointly fitting the power law constants and a monotonic function $f$ to
$$f\left(R\right)^{-\beta}=\left(\frac{N_c}N\right)^{\alpha_N}+\left(\frac{E_c}E\right)^{\alpha_E},$$
where $R$ is the mean episode return (or another performance metric such as TrueSkill), $N$ is the number of model parameters, and $E$ is the number of environment interactions. By also requiring the relationships between the constants from Lemma \ref{lemma} to hold, this provides us both with the power law constants, and with the desired function $f$ satisfying $f\left(R\right)=I$, where $I$ is intrinsic performance.

We perform this fit by using a black-box optimization algorithm such as CMA-ES to fit $\alpha_N$, $\alpha_E$ and $N_c$, which determine $\beta$ and $E_c$, with monotonic regression\footnote{\url{https://en.wikipedia.org/wiki/Isotonic_regression}} in the inner loop to fit $f$, using the squared error of the regression as the black-box loss function. We actually fit $\log\left(f\right)$ rather than $f$ in order to obtain a good fit to $I$ on a logarithmic scale, and we weight the data in proportion to $\frac 1E$ so that each interval is given equal weight on a logarithmic scale. In our Colab notebook, this routine is performed by the function \texttt{fit\_coeffs}.

This procedure seems to work well off-the-shelf, typically converging to a unique local minimum. However:

\begin{itemize}
\item When there is a lack of data or the data is very noisy, the local minimum may not be a global minimum, and the procedure can diverge to a degenerate solution.
\item It is necessary to first smooth learning curves so that they are mostly monotonic, to prevent the monotonic regression from overfitting. In our Colab notebook, we use the function \texttt{smooth}, which uses standard errors to automatically choose smoothing parameters (although note that we used slightly different smoothing parameters for MNIST).
\item As discussed in Section \ref{measurement-problem}, it is important to exclude data from early in training.
\end{itemize}

Our full procedure is therefore as follows.

\begin{itemize}
\item Smooth learning curves. Plot the smoothed curves on a logarithmic scale to check the monotonicity and fit, and adjust the smoothing parameters if necessary.
\item Exclude data from early in training, balancing the need for data against how much the early data skews the fit. Typically at least the first $\frac{1}{64}$ of training should be excluded.
\item Fit the power law constants and $f$ using the black-box optimization with monotonic regression routine.
\item Plot the fit to check the routine did not diverge. If it did, re-run routine, or constrain the constants and re-run, or include more data in step 2. If none of these fixes the divergence, then it may be necessary to collect more data.
\item Check the fit is not overly skewed by data from early in training. If it is, exclude more data in step 2.
\end{itemize}

This procedure led us to exclude the first 3 million environment interactions for Procgen, the first 2 billion environment interactions for Dota 2, and the first $2^{16}$, $2^{19}$ or $2^{22}$ environment interactions for MNIST depending on the period of training being considered, as discussed in Section \ref{variability-of-exponents-over-training}.

\subsection{Fitting to natural performance metrics}\label{fitting-to-natural-performance-metrics}

As discussed in Section \ref{natural-performance-metrics}, as well as fitting our power law with $I^{-\beta}$ on the left-hand side, as in equation \eqref{eq:intrinsic}, we also fit it using various other expressions, such as $e^{-\alpha_TT}$, where $T$ is TrueSkill and $\alpha_T$ is a fitted constant. When doing this, we adopt the convention that the constraints on $\beta$ and $E_c$ from Lemma \ref{lemma} should continue to hold. This necessitates introducing an additional multiplier, and instead fitting
$$T_ce^{-\alpha_TT}=\left(\frac{N_c}N\right)^{\alpha_N}+\left(\frac{E_c}E\right)^{\alpha_E}$$
for example, where $T_c$ is a fitted constant. Doing this allows us to continue interpret the left-hand side of this equation as $I^{-\beta}$.

To fit equations of this form, we continue use the same black-box optimization method, and simply replace the monotonic regression by another method of fitting $\log\left(f\right)$. For example, we may fit
$$f\left(T\right)^{-\beta}=T_ce^{-\alpha_TT}$$
by using linear regression to fit $\log\left(f\right)$. (Recall that $\beta$ is already determined by $\alpha_N$ and $\alpha_E$.)

The function from our Colab notebook, \texttt{fit\_coeffs}, provides options for fitting various functional forms for $f$, although it can sometimes be slow. (This is because it sometimes uses black-box optimization again in the inner loop for ease of implementation, even though this could be collapsed into the outer loop if speed were important.)

\newpage

\section{Hyperparameters}\label{appendix-hyperparameters}

Our default hyperparameters for Procgen, Dota 2 and MNIST are given in Tables \ref{hyperparameters-procgen}, \ref{hyperparameters-dota} and \ref{hyperparameters-mnist} respectively. We modified these defaults in two ways:

\begin{itemize}
\item
We adjusted the Adam step size as the model was scaled, as explained in Section \ref{learning-rates}.
\item
For Procgen and MNIST, we incorporated a batch ramp and learning rate schedule, as explained in Section \ref{batch-size}.
\end{itemize}

\begin{table}[h!]
\caption{Default PPG-EWMA hyperparameters for Procgen.}
\label{hyperparameters-procgen}
\centering
\begin{tabular}{@{}lll@{}}
\toprule
& Hyperparameter & Value \\
\midrule
PPO & Parallel environments & $1024$ \\
& Timesteps per rollout ($T$) & $256$ \\
& Minibatches per epoch & $8$ \\
& Adam step size ($\alpha$) & $5\times 10^{-4}$ \\
& Value function coefficient & $0.5$ \\
& Entropy coefficient & $0.01$ \\
& PPO clipping parameter ($\epsilon$) & Not used \\
& PPO KL penalty coefficient ($\beta$) & $1$ \\
& GAE discount rate ($\gamma$) & $0.999$ \\
& GAE bootstrapping parameter ($\lambda$) & $0.95$ \\
& Reward normalization? & Yes \\
& Advantage normalization? & Yes \\
& Total environment interactions & 200 million \\
\midrule
PPG & Policy iterations per phase ($N_\pi$) & $32$ \\
& Policy phase policy epochs ($E_\pi$) & $1$ \\
& Policy phase value function epochs ($E_V$) & $1$ \\
& Auxiliary phase epochs ($E_{\mathrm{aux}}$) & $6$ \\
& Auxiliary phase minibatches per epoch & $16N_\pi$ \\
& Auxiliary phase cloning coefficient ($\beta_{\mathrm{clone}}$) & $1$ \\
\midrule
PPG-EWMA & Proximal policy EWMA decay rate ($\beta_{\mathrm{prox}}$) & $\frac 89$ \\
\midrule
Batch ramp & Initial batch size multiplier & $\frac{1}{32}$ \\
\bottomrule
\end{tabular}
\end{table}

\begin{table}[h!]
\caption{PPO hyperparameters for Dota 2.}
\label{hyperparameters-dota}
\centering
\begin{tabular}{@{}ll@{}}
\toprule
Hyperparameter & Value \\
\midrule
Parallel environments & $6144$ \\
Timesteps per rollout ($T$) & $512$ \\
Minibatches per epoch & $32$ \\
Epochs ($E$) & $1$ \\
Adam step size ($\alpha$) & $10^{-4}$ to $10^{-3}$ \\
PPO clipping parameter ($\epsilon$) & $0.2$ \\
PPO KL penalty coefficient ($\beta$) & Not used \\
GAE bootstrapping parameter ($\lambda$) & $0.95$ \\
Total environment interactions & 13.6--82.6 billion \\
\bottomrule
\end{tabular}
\end{table}

\begin{table}[h!]
\caption{Default PPO-EWMA hyperparameters for MNIST in terms the horizon length $h$, which varied from 1 to 256.}
\label{hyperparameters-mnist}
\centering
\begin{tabular}{@{}lll@{}}
\toprule
& Hyperparameter & Value \\
\midrule
PPO & Parallel environments & $16$ \\
& Timesteps per rollout ($T$) & $512$ \\
& Minibatches per epoch & $8$ \\
& Epochs ($E$) & $1$ \\
& Adam step size ($\alpha$) & $1\times 10^{-3}$ \\
& Value function coefficient & $0.5$ \\
& Entropy coefficient & $0.01$ \\
& PPO clipping parameter ($\epsilon$) & Not used \\
& PPO KL penalty coefficient ($\beta$) & $1$ \\
& GAE discount rate ($\gamma$) & $1-\frac 2{h+1}$ \\
& GAE bootstrapping parameter ($\lambda$) & $1$ \\
& Reward normalization? & Yes \\
& Advantage normalization? & Yes \\
& Total environment interactions & $2^{25}$ \\
\midrule
PPO-EWMA & Proximal policy EWMA decay rate ($\beta_{\mathrm{prox}}$) & $\frac 89$ \\
\midrule
Batch ramp & Initial batch size multiplier & $\frac{\sqrt{h}}{64}$ \\
\bottomrule
\end{tabular}
\end{table}

\newpage

\subsection{Batch ramp and learning rate schedule}\label{batch-size}

As explained in Section \ref{learning-rates}, it was important to use a well-tuned learning rate schedule, and to use a schedule that works well for a variety of compute budgets. It was also important to use a batch ramp, i.e., to start with a small batch size and increase it over the course of training, because the critical batch size is smaller at the start of training, and we needed training to still be sample-efficient for small compute budgets. Without a batch ramp, we would have needed to adjust our power law, equation \eqref{eq:intrinsic}, in much the same way as the corresponding law for language \citep[equation (1.6)]{languagescalinglaws}, which uses $S_{\mathrm{min}}\left(S\right)$, the minimum number of optimization steps as estimated using a power law fit to the gradient noise scale.

Note, however, that increasing the batch size has a very similar effect to lowering the learning rate. To simplify matters, we used PPO-EWMA and PPG-EWMA, which are \textit{batch size-invariant} \citep{ppoewma}, allowing us to have almost the same effect as increasing the batch size by instead lowering the learning rate and increasing the center of mass of the proximal policy EWMA. We then considered only the batch size schedule, whether implemented explicitly or implicitly via these other hyperparameters.

To explore promising schedules, we implemented a greedy adaptive batch size algorithm, which tries doubling the batch size and switches if that performs better, or else backtracks and stays with the current batch size. We experimented with this on StarPilot's easy difficulty setting, using model sizes spanning a factor of around 2048. We found our algorithm to fairly consistently choose a schedule that can be well-approximated by the power law
\begin{equation*}
B=\max\left(B_{\mathrm{min}},\frac{E^{0.84}}{80}\right),
\end{equation*}
where $B$ is the batch size in interactions, $E$ is the total number of interactions so far, and $B_{\mathrm{min}}=256$ was our initial batch size.

Having fit this power law schedule on one Procgen environment, we tested it on several different Procgen environments, and found it to consistently outperform our usual fixed batch size both at the start and end of training. (Curiously, our schedule sometimes underperformed the fixed batch size in the middle of training. We believe this may be explained by the smaller initial batch size causing the entropy to fall too quickly at the start of training, highlighting a pitfall of the greedy approach.) In particular, we were able to use the same schedule on both the easy and hard difficulty settings. Our usual fixed batch size, on the other hand, was larger for the hard setting, corresponding to the fact that it was tuned to longer training runs.

The same schedule also worked well on our MNIST environment at every horizon length, although it was necessary to tune $B_{\mathrm{min}}$. Using too small a value for $B_{\mathrm{min}}$ seemed to result in an instability which could not always be recovered from. We found the optimal $B_{\mathrm{min}}$ to vary based on the horizon length $h$, and we took $B_{\mathrm{min}}=16\sqrt{h}$ (though taking $B_{\mathrm{min}}$ to have the form $A_0+A_1h$ would probably have made more theoretical sense in hindsight, given the results of Section \ref{effect-of-task-horizon-length}). If trying our schedule on other environments, we suggest tuning $B_{\mathrm{min}}$ to ensure stability at the start of training, but it is probably less important to tune the power law constants.

We used this batch size schedule for both our Procgen and MNIST experiments (although it would probably have been better to fully re-fit the schedule for MNIST). We implemented this using a batch size multiplier, explicitly reducing the batch size when the multiplier was less than 1, and changing the learning rate and center of mass of the proximal policy EWMA instead when the multiplier was greater than 1. With Procgen, for which we used PPG-EWMA, we also changed the number of policy iterations per phase, $N_\pi$, in proportion to the batch size, since we thought the number of optimization steps per phase should remain constant, and we rounded the batch size multiplier to the nearest power of two, with minimum and maximum multipliers of $\frac 1{32}$ and $4$ (corresponding to batch sizes of 1024 and 131072 respectively).

For Dota 2, we did not use a batch size schedule, since those experiments were carried out before we investigated batch size schedules.

\newpage

\section{Results in full}\label{appendix-results-in-full}

All the data from our experiments may be accessed using this Colab notebook: \url{https://colab.research.google.com/drive/1PzwZyXsi9jRdVCj1GJrS8JdOPBQ7LHZV}. This also includes code for analyzing this data, including model size and compute calculations, intrinsic performance and power law fitting, and generating all the plots in this paper.

\begin{figure}[H]
  \centerline{\scalebox{.55}{\input{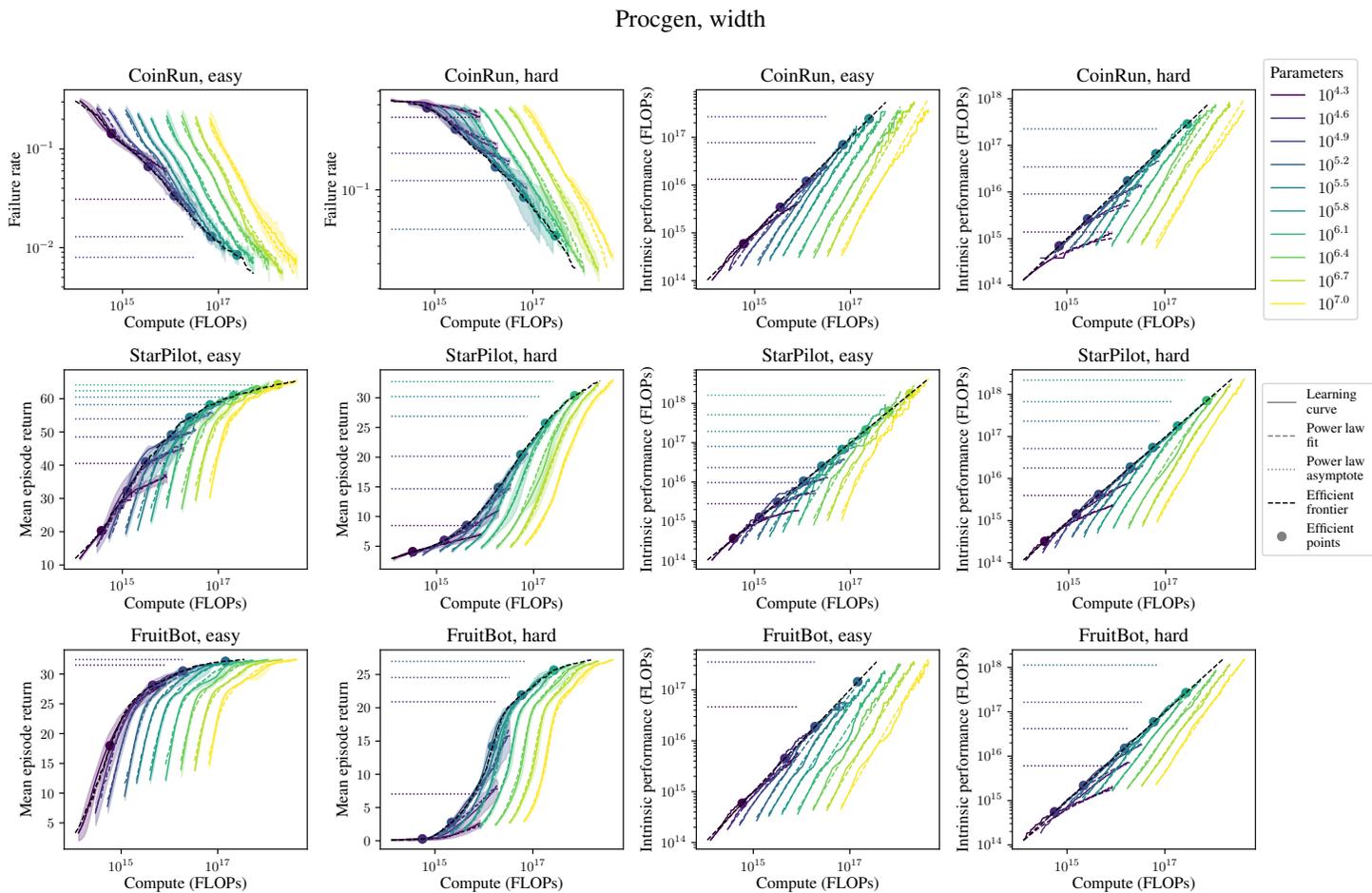}}} % PLACEHOLDER_procgen_fit_width.pgf
  \caption{Learning curves as a function of total training compute for our Procgen width-scaling experiments, together with their power law fits. Left half: mean episode return or failure rate, mean $\pm 1$ sample standard deviation over three seeds shown. Right half: intrinsic performance, mean only shown.}
  \label{figure-procgen-fit-width}
\end{figure}

Figures \ref{figure-procgen-fit-width}, \ref{figure-procgen-fit-depth}, \ref{figure-dota-fit} and \ref{figure-mnist-fit-late} show learning curves as a function of total training compute, together with their power law fits, for all of our experiments. On the left of each figure we show mean episode return (or failure rate for CoinRun and MNIST, or TrueSkill for Dota 2), with error bars showing mean $\pm 1$ sample standard deviation over the random seeds. On the right of each figure, we show intrinsic performance, with error bars hidden for clarity.

\begin{figure}[H]
  \centerline{\scalebox{.55}{\input{figures/procgen_fit_depth.pgf}}} % PLACEHOLDER_procgen_fit_depth.pgf
  \caption{Learning curves as a function of total training compute for our Procgen depth-scaling experiments, together with their power law fits. Left half: mean episode return or failure rate, mean $\pm 1$ sample standard deviation over three seeds shown. Right half: intrinsic performance, mean only shown.}
  \label{figure-procgen-fit-depth}
\end{figure}

\begin{figure}[H]
  \centerline{\scalebox{.75}{\input{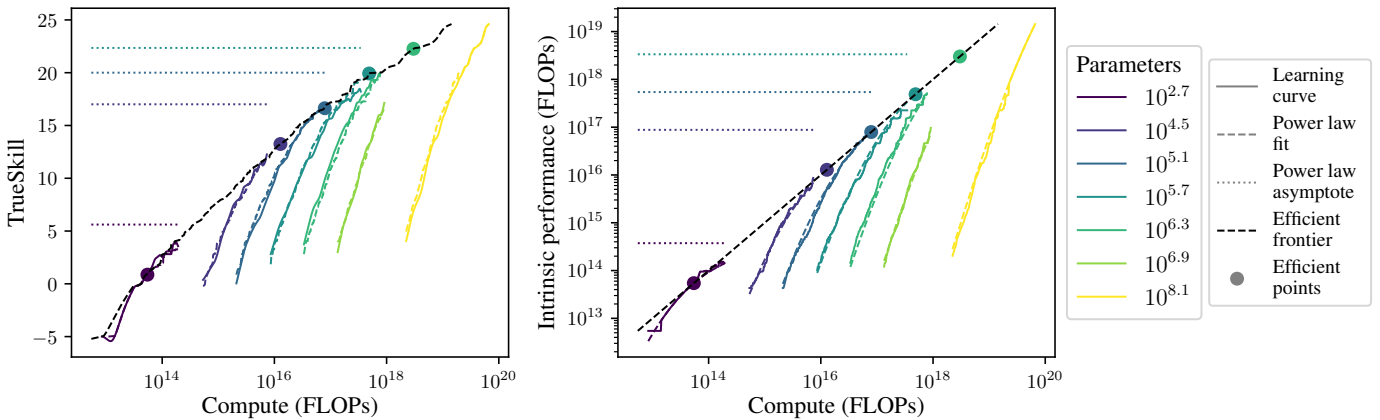}}}
  \caption{Learning curves as a function of total training compute for Dota 2, together with their power law fits. Only one random seed was used. Left: TrueSkill. Right: intrinsic performance.}
  \label{figure-dota-fit}
\end{figure}

\begin{figure}[H]
  \centerline{\scalebox{.55}{\input{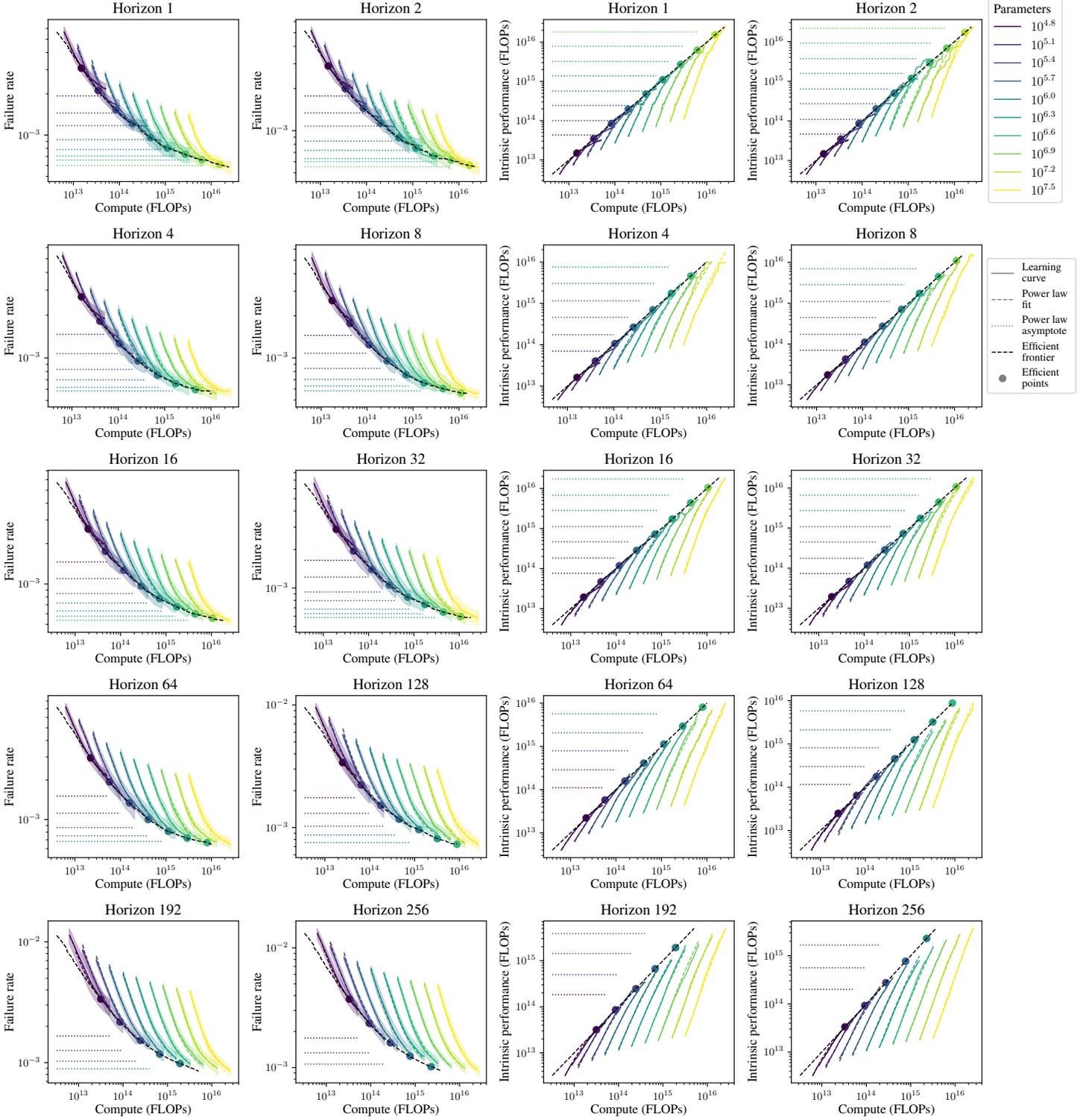}}} % PLACEHOLDER_mnist_fit_late.pgf
  \caption{Learning curves as a function of total training compute for MNIST, together with their power law fits, for the late period of training ($2^{22}$--$2^{25}$ environment interactions). Left half: failure rate, mean $\pm 1$ sample standard deviation over the middle-performing 16 of 20 random seeds shown. Right: intrinsic performance, mean only shown.}
  \label{figure-mnist-fit-late}
\end{figure}

\newpage

\section{Parameter and FLOP calculations}\label{appendix-parameter-and-flop-calculations}

In counting parameters and FLOPs, we apply the following principles:

\begin{itemize}
\item We only include the part of the network that is being scaled (ignoring things like embedding parameters), since we consider that to be the bottleneck.
\item We use round numbers (ignoring negligible contributions such as as biases and activations), for simplicity.
\item We include both rollout and optimization FLOPs (including any additional overhead of PPO-EWMA).
\item We treat an add-multiply as 2 FLOPs.
\end{itemize}

For example, we treat the forward pass of a dense layer as taking $2$ FLOPs per batch item per parameter, and a convolutional layer as taking $2h_{\mathrm{out}}w_{\mathrm{out}}$ FLOPs per batch item per parameter. We treat a backward pass as taking $2\times$ the FLOPs of a forward pass.

For the Procgen width-scaling experiments, we ignore the first convolution, since it scales as width (instead of as width squared), and has few parameters. Similarly, for the depth-scaling experiments, we ignore the final dense layer, since we only vary the number of convolutional layers. Unfortunately, as discussed in Section \ref{scaling-depth}, the final dense layer contains many parameters, which skews our constants. In both cases, we include both the policy and value networks, which are separate with identical architectures. We use PPG-EWMA with 1 policy epoch and 6 auxiliary epochs, totaling 9 forward and 7 backward passes per interaction.

For the Dota experiments, we ignore the embedding layer, considering only the LSTM. Since each interaction was used only once, we count 2 forward passes and 1 backward pass per interaction (1 forward pass for the rollout, and 1 forward-backward pass for optimization).

For the MNIST experiments, we ignore the first convolution, as for the Procgen width-scaling experiments. However, we only include the policy network, since the task of the value network is trivial (due to timesteps being independent). We use PPO-EWMA with 1 epoch, totaling 3 forward passes and 1 backward pass per interaction.

The numerical results of these calculations are as follows.

\begin{itemize}
\item\textbf{Procgen, scaling width}: for the width multiplier $w=2^{-3},2^{-2.5}2^{-2},\dots,2^{2.5}$, we count $1242112w^2$ parameters and $2652897280w^2$ FLOPs per interaction.
\item\textbf{Procgen, scaling depth}: for the number of residual blocks $b=1,2,4,\dots,64$, we count $5184b+1944$ parameters and $61046784b+81395712$ FLOPs per interaction.
\item\textbf{Dota 2}: for the LSTM size $s=8,64,128,256,512,1024,4096$, we count $8s^2$ parameters and $64s^2$ FLOPs per interaction.
\item\textbf{MNIST}: for the width multiplier $w=2^{-3},2^{-2.5}2^{-2},\dots,2^{2.5}$, we count $3948800w^2$ parameters and $95648000w^2$ FLOPs per interaction.
\end{itemize}

Note that one of our modeling assumptions is that the number of FLOPs per interaction is proportional to the number of parameters, but this is not true for our Procgen depth-scaling experiments. In other words, the number of FLOPs per param-interact, which is used to convert compute from units of parameters $\times$ interactions to units of FLOPs, is not constant. However, this number differs by at most 40\% from the mean of this number over the different depths, and so we simply used the mean when doing this conversion.

\newpage

\section{Fitted constants}\label{appendix-fitted-constants}

In this section we provide the constants $\alpha_N$, $\alpha_E$ and $N_c$, together with the values of $\beta$ and $E_c$ derived using Lemma \ref{lemma}, for our fitted power laws for intrinsic performance $I$ as given by equation \eqref{eq:intrinsic}. We also provide $I_{\mathrm{min}}$ and $I_{\mathrm{max}}$, the minimum and maximum intrinsic performance obtained during the span of interaction counts considered; our model is not able to predict mean episode return outside this range. Recall that the units of $I$ are parameters $\times$ interactions; the conversion to FLOPs may be performed using the values given in Appendix \ref{appendix-parameter-and-flop-calculations}.

We also provide the derived equations for optimal model size $N$ vs compute $C$ in PF-days. By substituting equation \eqref{eq:frontier} for the compute-efficient frontier into equation \eqref{eq:intrinsic}, these are given by
$$N=N_c\left(1+\frac{\alpha_N}{\alpha_E}\right)^{\frac 1{\alpha_N}}\left(\frac{C\times 10^{15}\times 24\times 3600}{\text{FLOPs per param-interact}}\right)^{\frac 1{1+\frac{\alpha_N}{\alpha_E}}}\quad\text{for}\quad N_{\mathrm{min}}\leq N\leq N_{\mathrm{max}}.$$
We take $N_{\mathrm{min}}$ and $N_{\mathrm{max}}$ to be the minimum and maximum model sizes we tested whose power law fit intersects the compute-efficient frontier somewhere between $I_{\mathrm{min}}$ and $I_{\mathrm{max}}$.

For our comparison to generative modeling, we use these equations for optimal model size $N$ vs compute $C$ in PF-days:

\begin{itemize}
\item Language \citep{chinchilla}: $N = (\frac C{1.4\times 10^{-18}})^{0.5}$
\item Language \citep{languagescalinglaws}: $N=(\frac C{3.3\times 10^{-13}})^{0.73}$
\item Image 32x32 \citep{scalingcompendium}: $N=(\frac C{1.6\times 10^{-13}})^{0.65}$
\end{itemize}

Further fitted constants, such as for single seeds, for different spans of interaction counts (see Section \ref{effect-of-task-horizon-length}), and fitted to natural performance metrics (see Section \ref{natural-performance-metrics}), may be found in this Colab notebook: \url{https://colab.research.google.com/drive/1PzwZyXsi9jRdVCj1GJrS8JdOPBQ7LHZV}.

\subsection{Procgen, scaling width}

The fitted constants for our Procgen width-scaling experiments are as follows.

\begin{table}[H]
\centering
\begin{tabular}{@{}llllllll@{}}
\toprule
Environment & $\alpha_N$ & $\alpha_E$ & $\beta$ & $N_c$ & $E_c$ & $I_{\mathrm{min}}$ & $I_{\mathrm{max}}$ \\
\midrule
CoinRun, easy & 0.542 & 0.462 & 0.249 & $2.53\times 10^{-2}$ & $2.49\times 10^{0}$ & $4.83\times 10^{10}$ & $2.55\times 10^{14}$ \\
CoinRun, hard & 0.759 & 0.576 & 0.328 & $1.55\times 10^{-1}$ & $8.00\times 10^{-1}$ & $6.07\times 10^{10}$ & $3.45\times 10^{14}$ \\
StarPilot, easy & 0.318 & 0.604 & 0.208 & $2.25\times 10^{-4}$ & $2.02\times 10^{2}$ & $4.88\times 10^{10}$ & $1.95\times 10^{15}$ \\
StarPilot, hard & 0.453 & 0.533 & 0.245 & $4.55\times 10^{-3}$ & $1.31\times 10^{1}$ & $5.43\times 10^{10}$ & $1.09\times 10^{15}$ \\
FruitBot, easy & 0.527 & 0.350 & 0.210 & $9.17\times 10^{-2}$ & $4.46\times 10^{-1}$ & $5.24\times 10^{10}$ & $1.67\times 10^{14}$ \\
FruitBot, hard & 0.478 & 0.346 & 0.201 & $1.14\times 10^{-1}$ & $2.96\times 10^{-1}$ & $6.00\times 10^{10}$ & $7.26\times 10^{14}$ \\
\bottomrule
\end{tabular}
\end{table}

These imply the following equations for optimal model size $N$ vs compute $C$ in PF-days.

\begin{itemize}
\item CoinRun, easy: $N=4.615\times 10^6\times C^{0.4600}$ for $19408\leq N\leq 310528$
\item CoinRun, hard: $N=6.881\times 10^6\times C^{0.4315}$ for $43668\leq N\leq 587092$
\item StarPilot, easy: $N=6.383\times 10^7\times C^{0.6549}$ for $19408\leq N\leq 4968448$
\item StarPilot, hard: $N=1.668\times 10^7\times C^{0.5404}$ for $19408\leq N\leq 1242112$
\item FruitBot, easy: $N=2.243\times 10^6\times C^{0.3994}$ for $19408\leq N\leq 174672$
\item FruitBot, hard: $N=6.631\times 10^6\times C^{0.4201}$ for $43668\leq N\leq 587092$
\end{itemize}

As discussed in Section \ref{natural-performance-metrics}, for CoinRun, we also fit power laws using the fail-to-success ratio $F$, excluding data for which $F>0.5$. As explained in Section \ref{fitting-to-natural-performance-metrics}, we replaced $I^{-\beta}$ with $\frac F{F_c}$, where $F_c$ is a fitted constant. The fitted constants for these power laws are as follows.

\begin{table}[H]
\centering
\begin{tabular}{@{}llllllll@{}}
\toprule
Difficulty & $\alpha_N$ & $\alpha_E$ & $\beta$ & $N_c$ & $E_c$ & $I_{\mathrm{min}}$ & $I_{\mathrm{max}}$ \\
\midrule
Easy & 0.899 & 1.007 & 0.475 & $1.00\times 10^{-2}$ & $2.33\times 10^{1}$ & $2.55\times 10^{10}$ & $2.60\times 10^{14}$ \\
Hard & 0.833 & 0.776 & 0.402 & $4.69\times 10^{-2}$ & $3.80\times 10^{0}$ & $5.14\times 10^{11}$ & $7.38\times 10^{14}$ \\
\bottomrule
\end{tabular}
\end{table}

\begin{table}[H]
\centering
\begin{tabular}{@{}ll@{}}
\toprule
Difficulty & $F_c$ \\
\midrule
Easy & $3.88\times 10^{4}$ \\
Hard & $2.52\times 10^{4}$ \\
\bottomrule
\end{tabular}
\end{table}

These imply the following relationships between $I$ and $F$.

\begin{itemize}
\item Easy: $I=4.57\times 10^9\times F^{-\frac 1{0.475}}$
\item Hard: $I=9.15\times 10^{10}\times F^{-\frac 1{0.402}}$
\end{itemize}

They also imply the following equations for optimal model size $N$ vs compute $C$ in PF-days.

\begin{itemize}
\item Easy: $N=1.216\times 10^7\times C^{0.5285}$ for $19408\leq N\leq 587092$
\item Hard: $N=1.148\times 10^7\times C^{0.4822}$ for $77632\leq N\leq 1242112$
\end{itemize}

\subsection{Procgen, scaling depth}

The fitted constants for our Procgen depth-scaling experiments are as follows.

\begin{table}[H]
\centering
\begin{tabular}{@{}llllllll@{}}
\toprule
Environment & $\alpha_N$ & $\alpha_E$ & $\beta$ & $N_c$ & $E_c$ & $I_{\mathrm{min}}$ & $I_{\mathrm{max}}$ \\
\midrule
CoinRun, easy & 0.351 & 0.469 & 0.201 & $2.64\times 10^{-4}$ & $1.26\times 10^{2}$ & $5.43\times 10^{9}$ & $3.72\times 10^{13}$ \\
CoinRun, hard & 0.336 & 0.581 & 0.213 & $1.02\times 10^{-4}$ & $4.47\times 10^{2}$ & $6.58\times 10^{9}$ & $6.24\times 10^{13}$ \\
StarPilot, easy & 0.800 & 0.821 & 0.405 & $9.65\times 10^{-3}$ & $1.87\times 10^{1}$ & $1.70\times 10^{10}$ & $5.52\times 10^{13}$ \\
StarPilot, hard & 0.380 & 0.381 & 0.190 & $2.87\times 10^{-3}$ & $9.11\times 10^{0}$ & $1.58\times 10^{10}$ & $5.21\times 10^{13}$ \\
FruitBot, easy & 0.539 & 0.564 & 0.276 & $2.92\times 10^{-3}$ & $2.77\times 10^{1}$ & $9.58\times 10^{9}$ & $3.76\times 10^{13}$ \\
FruitBot, hard & 0.401 & 0.463 & 0.215 & $1.23\times 10^{-3}$ & $3.26\times 10^{1}$ & $1.34\times 10^{10}$ & $4.64\times 10^{13}$ \\
\bottomrule
\end{tabular}
\end{table}

These imply the following equations for optimal model size $N$ vs compute $C$ in PF-days. Note, however, that:

\begin{itemize}
\item As discussed in Section \ref{scaling-depth}, we exclude the final dense layer, which would have accounted for between 16\% and 90\% of the parameters, depending on the depth. This skews the leading constants here.
\item As discussed in Appendix \ref{appendix-parameter-and-flop-calculations}, we also ignored the variation in the number of FLOPs per param-interact between models of different depths, leading to errors of up to 40\%.
\end{itemize}

\begin{itemize}
\item CoinRun, easy: $N=1.390\times 10^6\times C^{0.5723}$ for $7128\leq N\leq 43416$
\item CoinRun, hard: $N=3.962\times 10^6\times C^{0.6337}$ for $7128\leq N\leq 167832$
\item StarPilot, easy: $N=2.202\times 10^6\times C^{0.5063}$ for $7128\leq N\leq 167832$
\item StarPilot, hard: $N=1.410\times 10^6\times C^{0.5007}$ for $7128\leq N\leq 84888$
\item FruitBot, easy: $N=1.172\times 10^6\times C^{0.5110}$ for $7128\leq N\leq 84888$
\item FruitBot, hard: $N=1.671\times 10^6\times C^{0.5359}$ for $7128\leq N\leq 84888$
\end{itemize}

\subsection{Dota 2}

As explained in Sections \ref{natural-performance-metrics} and \ref{fitting-to-natural-performance-metrics}, we fit power laws to $I^{-\beta}$, $T_ce^{-\alpha_TT}$, $T_c\left(e^{-\alpha_TT}-e^{\alpha_TT^\ast}\right)$ and $T_c\left(T^\ast-T\right)^{\alpha_T}$, where $I$ is intrinsic performance, $T$ is TrueSkill, and $\alpha_T$, $T_c$ and $T^\ast$ are fitted constants. The fitted constants for these different functional forms are as follows.

\begin{table}[H]
\centering
\begin{tabular}{@{}llllllll@{}}
\toprule
Fit to & $\alpha_N$ & $\alpha_E$ & $\beta$ & $N_c$ & $E_c$ & $I_{\mathrm{min}}$ & $I_{\mathrm{max}}$ \\
\midrule
$I^{-\beta}$ & 0.186 & 0.593 & 0.141 & $1.98\times 10^{-8}$ & $1.04\times 10^{6}$ & $6.83\times 10^{11}$ & $1.79\times 10^{18}$ \\
$T_ce^{-\alpha_TT}$ & 0.180 & 0.486 & 0.131 & $3.53\times 10^{-8}$ & $3.33\times 10^{5}$ & $4.62\times 10^{11}$ & $2.24\times 10^{17}$ \\
$T_c(e^{-\alpha_TT}-e^{\alpha_TT^\ast})$ & 0.181 & 0.560 & 0.137 & $2.07\times 10^{-8}$ & $8.32\times 10^{5}$ & $6.31\times 10^{11}$ & $1.77\times 10^{18}$ \\
$T_c(T^\ast-T)^{\alpha_T}$ & 0.183 & 0.569 & 0.138 & $2.06\times 10^{-8}$ & $8.82\times 10^{05}$ & $6.71\times 10^{11}$ & $1.23\times 10^{18}$ \\
\bottomrule
\end{tabular}
\end{table}

\begin{table}[H]
\centering
\begin{tabular}{@{}llll@{}}
\toprule
Fit to & $\alpha_T$ & $T_c$ & $T^\ast$ \\
\midrule
$I^{-\beta}$ & - & - & - \\
$T_ce^{-\alpha_TT}$ & 0.0572 & $2.16\times 10^{-2}$ & - \\
$T_c(e^{-\alpha_TT}-e^{\alpha_TT^\ast})$ & 0.0402 & $2.40\times 10^{-2}$ & 35.43 \\
$T_c(T^\ast-T)^{\alpha_T}$ & 2.84 & $2.14\times 10^{-7}$ & 54.01 \\
\bottomrule
\end{tabular}
\end{table}

As discussed in Section \ref{natural-performance-metrics}, we have less confidence in the last two functional forms, which is reflected in the very different estimates for $T^\ast$, which represents the maximum attainable TrueSkill for the family of models we trained.

These imply the following relationships between $I$ and $T$ for the last three fits.

\begin{itemize}
\item \makebox[3.25cm][l]{$T_ce^{-\alpha_TT}$:} $I=4.93\times 10^{12}\times 1.5462^T$
\item \makebox[3.25cm][l]{$T_c(e^{-\alpha_TT}-e^{\alpha_TT^\ast})$:} $I=6.49\times 10^{11}\times\left(1.0410^{-T}-1.0410^{-35.43}\right)^{-\frac 1{0.137}}$
\item \makebox[3.25cm][l]{$T_c(T^\ast-T)^{\alpha_T}$:} $I=1.48\times 10^{48}\times\left(54.01-T\right)^{-\frac{2.84}{0.138}}$
\end{itemize}

They also imply the following equations for optimal model size $N$ vs compute $C$ in PF-days.

\begin{itemize}
\item \makebox[3.25cm][l]{$I^{-\beta}$:} $N=2.703\times 10^7\times C^{0.7617}$ for $512\leq N\leq 2097152$
\item \makebox[3.25cm][l]{$T_ce^{-\alpha_TT}$:} $N=1.607\times 10^7\times C^{0.7302}$ for $512\leq N\leq 524288$
\item \makebox[3.25cm][l]{$T_c(e^{-\alpha_TT}-e^{\alpha_TT^\ast})$:} $N=2.305\times 10^7\times C^{0.7552}$ for $512\leq N\leq 2097152$
\item \makebox[3.25cm][l]{$T_c(T^\ast-T)^{\alpha_T}$:} $N=2.385\times 10^7\times C^{0.7567}$ for $512\leq N\leq 2097152$
\end{itemize}

\subsection{MNIST}

The fitted constants for our MNIST experiments are as follows. As discussed in Section \ref{variability-of-exponents-over-training}, these constants are for the late period of training ($2^{22}$--$2^{25}$ environment interactions). Recall also that the horizon $h$ is such that the interval $\left[0,h-1\right]$ has the same center of mass as an exponentially-weighted moving average with decay parameter $\gamma$, i.e., $\gamma=1-\frac 2{h+1}$.

\begin{table}[H]
\centering
\begin{tabular}{@{}llllllll@{}}
\toprule
Horizon & $\alpha_N$ & $\alpha_E$ & $\beta$ & $N_c$ & $E_c$ & $I_{\mathrm{min}}$ & $I_{\mathrm{max}}$ \\
\midrule
1 & 0.263 & 1.050 & 0.210 & $9.79\times 10^{-6}$ & $9.43\times 10^{3}$ & $1.79\times 10^{11}$ & $1.00\times 10^{15}$ \\
2 & 0.265 & 0.979 & 0.208 & $1.32\times 10^{-5}$ & $6.30\times 10^{3}$ & $1.87\times 10^{11}$ & $9.66\times 10^{14}$ \\
4 & 0.284 & 0.791 & 0.209 & $4.21\times 10^{-5}$ & $1.50\times 10^{3}$ & $1.94\times 10^{11}$ & $4.19\times 10^{14}$ \\
8 & 0.276 & 0.826 & 0.207 & $2.83\times 10^{-5}$ & $2.33\times 10^{3}$ & $1.80\times 10^{11}$ & $6.24\times 10^{14}$ \\
16 & 0.252 & 0.830 & 0.193 & $1.59\times 10^{-5}$ & $3.78\times 10^{3}$ & $1.62\times 10^{11}$ & $7.69\times 10^{14}$ \\
32 & 0.263 & 0.856 & 0.201 & $1.73\times 10^{-5}$ & $3.83\times 10^{3}$ & $1.59\times 10^{11}$ & $7.47\times 10^{14}$ \\
64 & 0.307 & 0.736 & 0.217 & $7.27\times 10^{-5}$ & $8.40\times 10^{2}$ & $1.64\times 10^{11}$ & $4.16\times 10^{14}$ \\
128 & 0.315 & 0.769 & 0.224 & $6.27\times 10^{-5}$ & $1.08\times 10^{3}$ & $1.45\times 10^{11}$ & $3.64\times 10^{14}$ \\
192 & 0.330 & 0.688 & 0.223 & $1.22\times 10^{-4}$ & $4.86\times 10^{2}$ & $1.33\times 10^{11}$ & $2.08\times 10^{14}$ \\
256 & 0.358 & 0.681 & 0.235 & $2.11\times 10^{-4}$ & $3.04\times 10^{2}$ & $1.33\times 10^{11}$ & $1.53\times 10^{14}$ \\
\bottomrule
\end{tabular}
\end{table}

These imply the following equations for optimal model size $N$ vs compute $C$ in PF-days.

\begin{itemize}
\item Horizon 1:\hphantom{00} $N=1.586\times 10^{10}\times C^{0.7999}$ for $61700\leq N\leq 15795200$
\item Horizon 2:\hphantom{00} $N=1.309\times 10^{10}\times C^{0.7871}$ for $61700\leq N\leq 15795200$
\item Horizon 4:\hphantom{00} $N=5.507\times 10^9\times C^{0.7357}$ for $61700\leq N\leq 3948800$
\item Horizon 8:\hphantom{00} $N=6.406\times 10^9\times C^{0.7493}$ for $61700\leq N\leq 7739648$
\item Horizon 16:\hphantom{0} $N=7.787\times 10^9\times C^{0.7671}$ for $61700\leq N\leq 7739648$
\item Horizon 32:\hphantom{0} $N=7.535\times 10^9\times C^{0.7652}$ for $61700\leq N\leq 7739648$
\item Horizon 64:\hphantom{0} $N=2.746\times 10^9\times C^{0.7053}$ for $61700\leq N\leq 3948800$
\item Horizon 128: $N=2.681\times 10^9\times C^{0.7092}$ for $61700\leq N\leq 3948800$
\item Horizon 192: $N=1.376\times 10^9\times C^{0.6757}$ for $61700\leq N\leq 987200$
\item Horizon 256: $N=9.876\times 10^8\times C^{0.6553}$ for $61700\leq N\leq 987200$
\end{itemize}

\newpage

\section{Proof of the lemma}\label{appendix-proof-of-the-lemma}

\begin{proof}[Proof of Lemma \ref{lemma}]
We may write $I\left(N,E\right)$ as a function of $N$ and compute $C:=NE$:
$$I\left(N,C\right)^{-\beta}=\left(\frac{N_c}N\right)^{\alpha_N}+\left(\frac{E_cN}{C}\right)^{\alpha_E}.$$

The compute-efficient frontier is defined by the value of $N$ that maximizes $I\left(N,C\right)$ for each $C$. Equivalently, since $\beta>0$, this value of $N$ minimizes $I\left(N,C\right)^{-\beta}$, and so it satisfies
$$\frac\partial{\partial N}\left(I\left(N, C\right)^{-\beta}\right)=0.$$
Differentiating and multiplying through by $N$, this equation becomes
$$-\alpha_N\left(\frac{N_c}N\right)^{\alpha_N}+\alpha_E\left(\frac{E_cN}{C}\right)^{\alpha_E}=0.$$
Eliminating $C$, this is exactly equation \eqref{eq:frontier}, as required.

By assumption, we also have $I\left(N,E\right)=NE$ along the compute-efficient frontier. Substituting \eqref{eq:frontier} into $I\left(N,E\right)$, this equation becomes
\begin{equation}\label{eq:frontier-alternative}
\left(1+\frac{\alpha_N}{\alpha_E}\right)\left(\frac{N_c}N\right)^{\alpha_N}=\left(NE\right)^{-\beta}.
\end{equation}
Thus both equations \eqref{eq:frontier} and \eqref{eq:frontier-alternative} are power law relationships between $N$ and $E$ that hold along the compute-efficient frontier, so we may simply equate exponents and constants. Equating exponents,
$$\frac{\alpha_N}{\alpha_E}=\frac{\alpha_N}{\beta} - 1\qquad\text{and hence}\qquad\frac 1\beta=\frac 1{\alpha_N}+\frac 1{\alpha_E},$$
as required. Equating constants,
$$\left(\frac{\alpha_N}{\alpha_E}\right)^{\frac 1{\alpha_E}}N_c^{\frac{\alpha_N}{\alpha_E}}E_c^{-1}=\left(1+\frac{\alpha_N}{\alpha_E}\right)^{\frac 1\beta}N_c^{\frac{\alpha_N}\beta},$$
and hence
$$\frac 1{N_cE_c}=\left(1+{\frac{\alpha_N}{\alpha_E}}\right)^{\frac 1{\alpha_N}+\frac 1{\alpha_E}}\left(\frac{\alpha_E}{\alpha_N}\right)^{\frac 1{\alpha_E}}=\left(1+\frac{\alpha_N}{\alpha_E}\right)^{\frac 1{\alpha_N}}\left(1+\frac{\alpha_E}{\alpha_N}\right)^{\frac 1{\alpha_E}},$$
as required.
\end{proof}

\newpage

\section{Proof sketch of the proposition}\label{appendix-proof-sketch-of-the-proposition}

A formal statement and proof of Proposition \ref{proposition} would require a formal analysis of Vanilla Policy Gradient, which is beyond the scope of this work. Instead, we provide a proof sketch in which we make approximations informally.

\begin{proof}[Proof sketch of Proposition \ref{proposition}]
The horizon length $h$ only affects the algorithm via GAE, which in the case $\lambda=1$ produces the value function targets and advantage estimates
$$\hat V_t:=r_t+\gamma r_{t+1}+\dots+\gamma^{T-t} r_T=r_t+\gamma\hat V_{t+1}\qquad\text{and}$$
$$\hat A_t:=\hat V_t-V\left(s_t\right)=r_t-V\left(s_t\right)+\gamma\hat V_{t+1},\qquad\quad$$
where $V$ is the value function. Since timesteps are independent, $\gamma\hat V_{t+1}$ is independent of $s_t$ and $a_t$, and so should be thought of as noise. The value function will quickly learn to incorporate the mean of this noise, and so
$$V\left(s_t\right)\approx V^0\left(s_t\right)+\mathbb E\left[\gamma\hat V_{t+1}\right],$$
where $V^0\left(s_t\right)$ is the ``immediate reward value function'' that would have been obtained had we used the value function targets $\hat V^0_t:=\hat V_t-\mathbb E\left[\gamma\hat V_{t+1}\right]$. Writing $\epsilon:=\gamma\hat V_{t+1}-\mathbb E\left[\gamma\hat V_{t+1}\right]$ for the zero-mean component of $\gamma\hat V_{t+1}$, we obtain
$$\hat V^0_t=r_t+\epsilon\qquad\text{and}$$
$$\hat A_t\approx r_t-V^0\left(s_t\right)+\epsilon.\qquad\quad$$
In other words, the entire impact of varying $h$ is that it changes the variance of the noise term $\epsilon$ added to the value function targets and advantage estimates.

Let us now analyze the policy gradient, which equals
$$\hat{\mathbb E}_t\left[\nabla_\theta \rho_t\left(\theta\right)\hat A_t\right]\approx\hat{\mathbb E}_t\left[\nabla_\theta \rho_t\left(\theta\right)\left(r_t-V^0\left(s_t\right)+\epsilon\right)\right],$$
where $\rho_t\left(\theta\right):=\frac{\pi_\theta\left(a_t|s_t\right)}{\pi_{\theta_{\mathrm{old}}}\left(a_t| s_t\right)}$. Since $\epsilon$ is independent of $s_t$ and $a_t$ and $\mathbb E\left[\epsilon\right]=0$, the covariance matrix of this decomposes as
$$\boldsymbol\Sigma_\theta+\boldsymbol\Phi_\theta\mathrm{Var}\left[\epsilon\right],$$
where $\boldsymbol\Sigma_\theta$ is the covariance matrix of $\nabla_\theta \rho_t\left(\theta\right)\left(r_t-V^0\left(s_t\right)\right)$, and $\boldsymbol\Phi_\theta:=\mathbb E\left[\nabla_\theta\rho_t\left(\theta\right)\nabla_\theta^{\mathsf T}\rho_t\left(\theta\right)\right]$ is the uncentered covariance matrix of $\nabla_\theta\rho_t\left(\theta\right)$.

Note that $V^0\left(s_t\right)$ simply estimates $\mathbb E\left[r_t\right]$, which does not depend on $h$. The variance of $V^0\left(s_t\right)$ does depend on $h$ via the addition of $\epsilon$ to the value function targets, but this additional variance is small compared to the variance of $\epsilon$ itself. We may therefore treat $\boldsymbol\Sigma_\theta$ as approximately independent of $h$. 

It remains to express $\mathrm{Var}\left[\epsilon\right]$ in terms of $h$. We assume that $T$ is large enough compared to $h$ that we may take $T\to\infty$. (In our experiments, we use rollouts of length 512 and $h\leq 256$.) Thus
$$
\begin{aligned}
\mathrm{Var}\left[\epsilon\right]&=\mathrm{Var}\left[\gamma\hat V_{t+1}\right]\\
&=\left(\gamma^2+\gamma^4+\gamma^6+\dots\right)\mathrm{Var}\left[r_t\right]\\
&=\frac{\gamma^2}{1-\gamma^2}\mathrm{Var}\left[r_t\right]\\
&=\frac 14\left(h+\frac 1h-2\right)\mathrm{Var}\left[r_t\right].
\end{aligned}
$$
Hence the covariance matrix of the policy gradient is approximately
$$\boldsymbol\Sigma_\theta+\boldsymbol\Pi_\theta\left(h+\frac 1h-2\right),$$
where $\boldsymbol\Sigma_\theta$ and $\boldsymbol\Pi_\theta:=\frac 14\mathrm{Var}\left[r_t\right]\boldsymbol\Phi_\theta$ are symmetric positive semi-definite matrices that do not depend on $h$, as required.
\end{proof}

\end{document}